\documentclass[twoside]{article}
\usepackage[numbers]{natbib}
\usepackage{aistats2025}
\usepackage{amsmath}
\usepackage{hyperref}
\usepackage{graphicx}
\usepackage{xcolor}
\usepackage{amssymb}
\usepackage{multirow}
\usepackage{makecell}
\usepackage{url}
\usepackage{booktabs}
\newtheorem{definition}{Definition}[section]
\newtheorem{proposition}{Proposition}[section]
\newtheorem{assumption}{Assumption}[section]
\newtheorem{theorem}{Theorem}[section]
\newtheorem{remark}{Remark}[section]

\allowdisplaybreaks

\begin{document}

%

%

\twocolumn[

\aistatstitle{A Theoretical Understanding of Chain-of-Thought: Coherent Reasoning and Error-Aware Demonstration}

\aistatsauthor{ $\text{Yingqian Cui}$ \And $\text{Pengfei He}$ \And  $\text{Xianfeng Tang}$ \\ }
\aistatsaddress{ Michigan State University \And Michigan State University \And Amazon } 
\aistatsauthor{$\text{Qi He}$ \And $\text{Chen Luo}$ \And $\text{Jiliang Tang}$ \And $\text{Yue Xing}$ }
\aistatsaddress{ Amazon\And Amazon \And Michigan State University \And Michigan State University } ]

\begin{abstract}
    Few-shot Chain-of-Thought (CoT) prompting has demonstrated strong performance in improving the reasoning capabilities of large language models (LLMs).
    While theoretical investigations have been conducted to understand CoT, 
    the underlying transformer used in these studies isolates the CoT reasoning process into separated in-context learning steps (Stepwise ICL).
    In this work, we theoretically show that, compared to Stepwise ICL, the transformer gains better error correction ability and more accurate predictions if the reasoning from earlier steps (Coherent CoT) is integrated.
    Given that this coherent reasoning changes the behavior of the transformer, we further investigate the sensitivity of the transformer with Coherent CoT when the demonstration examples are corrupted at the inference stage. Our theoretical results indicate that the transformer is more sensitive to errors in intermediate reasoning steps than the final outcome. Building upon this observation, we propose an improvement on CoT by incorporating both correct and incorrect reasoning paths in the demonstration.  
    Our experiments validate the effectiveness of the proposed approach.
\end{abstract}

\section{Introduction}
Few-shot Chain-of-Thought (CoT) prompting has emerged as a highly effective technique to enhance the reasoning capabilities of large language models (LLMs)~\citep{wei2022chain}. Given a few examples of step-by-step reasoning at the \textit{inference stage},
the model can generalize the reasoning process to new tasks, demonstrating significant improvement in solving complex problems, particularly in mathematical reasoning and commonsense inference~\citep{wei2022chain,fu2023chain,lyu2023faithful}.

Aside from these empirical successes, recent efforts have provided valuable theoretical insights into CoT. In the literature, two main approaches are typically used to analyze the ability of transformers in CoT tasks (and more broadly, ICL). In the first approach, to show the existence of transformers that are capable of performing CoT, people explicitly construct a transformer by specifying its parameters and assigning each component to perform a specific task. The second approach defines a specific data format to organize different reasoning steps and then trains a transformer model to learn a step-by-step prediction process. Subsequent analysis focuses on the properties of this trained model. Using these approaches, previous studies are able to explain the expressiveness power of the transformer in CoT tasks and understand how different components of the model contribute to the multi-step reasoning~\citep{li2023dissecting, li2024chain,feng2023towards,tutunov2023can}.

However, a key limitation of these analyses is that they often overlook the connections among multiple reasoning steps. In the approach where a transformer is directly constructed to perform CoT, predictions at each reasoning step are based solely on the result from the previous step. Similarly, when training a transformer from scratch on CoT tasks, to simplify the analysis, a ``filtering" process is used to retain only the information from the most recent step
~\cite{li2023dissecting}. In summary, when constructing/training the transformer from either way mentioned above, the prediction process for a training sample (a CoT prompt) involves multiple reasoning steps, and each step only focuses on the immediate task at hand without incorporating earlier reasoning steps. We refer to this prediction process as ``Stepwise ICL" (formally defined in Definition \ref{def:icl}).




However, in real-world scenarios, e.g., next token prediction, the LLM takes into account all the previous tokens in the context window, rather than treating each step in isolation (referred to as ``Coherent CoT", defined in Definition \ref{def:cot}).
Observing this discrepancy between Stepwise ICL and Coherent CoT, we extend the theoretical framework in \cite{zhang2023trained} and study the properties of a model trained using Coherent CoT and compare it with the model trained with Stepwise ICL.
Based on our result, using Coherent CoT instead of Stepwise ICL during the \textit{training stage} provides better prediction performance. Intuitively, when treating CoT as a holistic process -- where later steps integrate the reasoning from earlier steps -- the transformer will consider the potential errors in previous predictions and adjust subsequent predictions accordingly, which provides a form of self-correction and enhances the prediction performance. 

While Coherent CoT potentially outperforms Stepwise ICL as elaborated in our first contribution, it requires training the model on the entire reasoning chain, which alters the model's optimal parameters and changes the behavior of the model. 
{This change introduces uncertainty about which steps in the reasoning chain are most sensitive to errors, highlighting the necessity for sensitivity analysis. }
This leads to our \textbf{second contribution}, which focuses on the sensitivity of the trained Coherent CoT model to perturbations in the demonstration examples at the \textit{inference stage}. To quantify the sensitivity, we examine how the Coherent CoT model reacts with random perturbations at different reasoning steps. We reveal that, during inference, the Coherent CoT model is more sensitive to {noise} in the intermediate reasoning steps of the demonstration examples than to inaccuracies in their final outcomes.

Inspired by the sensitivity analysis, our \textbf{third contribution} is to propose a prompt composing method to enhance CoT performance at the \textit{inference stage}. 
Based on our sensitivity result, CoT is more sensitive to possible incorrectness at the intermediate reasoning steps, thus improving the accuracy of these steps can better enhance the overall CoT performance.
We propose to incorporate both correct and incorrect reasoning paths in the demonstrations to enhance the accuracy of the intermediate reasoning steps. Experiments are conducted to validate the effectiveness of the proposed method.


\section{Related Works}
\paragraph{Empirical Insights in CoT.}
Chain-of-Thought (CoT) prompting, introduced by \cite{wei2022chain}, has proven to be highly effective in enhancing the reasoning capabilities of LLMs by breaking complex tasks into step-by-step processes. This technique has been expanded into various variants and extensions, including Zero-shot CoT~\citep{kojima2022large}, Self-Consistency~\citep{wang2022self}, Auto-CoT~\citep{zhang2022automatic}, Tree-of-Thought~\citep{yao2023tree}, and Graph-of-Thought~\cite{besta2023graph}, to further improve efficiency or model performance.

Building upon the success of CoT, recent studies seek to deepen the understanding behind CoT by empirically exploring its mechanisms. For example, \cite{wu2023analyzing} finds that CoT enables the model to maintain the attention robust and stable on the semantically relevant tokens in the prompt. 
\cite{madaan2022text} defines the key components of a prompt as symbols, patterns, and text, and investigates how each of these elements and their interaction contribute to the superior performance of CoT.
Additionally, \cite{wang2022towards} validates that the relevance of the demonstration example to the query and the correct ordering of reasoning steps are key factors for the effectiveness of CoT, and \cite{jin2024impact} examines the relationship between CoT's effectiveness and reasoning step length. Based on \cite{jin2024impact}, simpler tasks require fewer reasoning steps, while more complex tasks benefit greatly from more detailed inference sequences.

\paragraph{Theories in ICL and CoT.} 
The mechanism of ICL has been extensively studied in the theoretical literature. For example, studies such as \cite{von2023transformers, ahn2023transformers, akyurek2022learning, zhang2023trained,huang2023context} have explained how ICL learns to perform linear regression using gradient descent. The work by \cite{cheng2023transformers} extends these analyses by investigating how transformers can apply ICL to non-linear functions, while \cite{bai2023transformers} focuses on generalized linear models, ridge regression, and LASSO. Additionally, \cite{cui2024superiority,chen2024training} explains why multi-head attention is preferred than single-head attention when performing ICL.

Besides ICL, recent research also starts to establish theoretical frameworks to understand CoT. 
For example, Li et al.\cite{li2023dissecting} offers a specific framework that views CoT as a series of ICL components, each addressing a smaller subproblem. \cite{feng2023towards} constructs a transformer that solves arithmetic and linear equation tasks using CoT. 
\cite{prystawski2024think} offers a Bayesian perspective on how intermediate steps improve reasoning. Additionally, \cite{li2024chain} provides a $\text{TC}^\circ$ upper bound for the expressiveness of constant-precision transformer, and \cite{tutunov2023can} introduces a two-level hierarchical graphical model
to explain how LLMs generate sequences of reasoning steps.

\section{Theoretical Results}

We extend the existing theory of ICL in transformers, e.g., \cite{zhang2023trained}, to a CoT scenario to conduct our theoretical investigation. Briefly speaking, there are two key observations: (1) 
Compared to Stepwise ICL, using Coherent CoT at the training stage {results in a model with} better inference performance.
(2) When noises exist in the demonstration examples at the inference stage, the model with Coherent CoT is more sensitive to perturbations in the reasoning steps than the inaccuracies in the final response.

In the following, we introduce some setups in Section \ref{sec:model_setup} for data generation and the transformer architecture, then present the theoretical results for (1) and (2) respectively in Section \ref{sec:theory} and \ref{sec:sensitivity}.

\subsection{Model Setup}\label{sec:model_setup}

\paragraph{Data generation {process}.} 
{Since demonstration examples are needed in the prompt in CoT, we}
define how the examples as well as the query data are generated as follows.

\begin{assumption} [Data Generation Process]
\label{ass:gen} In each prompt, the examples $(x_i,z_i, y_i)$ and the query data $(x_q,z_q, y_q)$ are i.i.d sampled from the following ``two-layer" noisy regression:
    \begin{itemize}\vspace{-0.1in}
    \item The {independent variable} $x \in \mathbb{R}^d \in N(0,I_d)$.\vspace{-0.1in}
    \item The {intermediate response} $z=\beta^\top x$.\vspace{-0.1in}
    \item The {final response} $y=z+ \epsilon $, $\epsilon \in N (0, \sigma^2)$.\vspace{-0.1in}
    \end{itemize}
For each prompt, all the examples and the query data share the same $\beta$. In different prompts, $\beta$ is i.i.d. uniformly sampled from unit sphere, i.e., $\|\beta\|_2^2=1$.
\end{assumption}

Assumption \ref{ass:gen} is primarily based on \cite{li2023dissecting} with some modifications  about the relation between $y$ and $z$. Following \cite{li2023dissecting}, we assume $x$ follows a Gaussian distribution to simplify the analysis. In terms of the final response $y$, we assume $y = \beta^\top x + \epsilon$, a linear mapping of $x$ with added noise. To improve the prediction of $y$, an intermediate response $z = \beta^\top x$  is introduced. As a noise-free representation of the relationship between $x$ and $y$, introducing $z$ helps to mitigate the impact of noise $\epsilon$ and guide a more accurate prediction of $y$. {An additional discussion on potential relaxations of the assumptions can be found in Remark \ref{rem:gaussian}.}

\paragraph{Coherent CoT and Stepwise ICL.}
We define ``Coherent CoT" and ``Stepwise ICL" as follows:
\begin{definition}[Stepwise ICL]\label{def:icl}
There are two separate ICL steps involved in the Stepwise ICL process, and each is performed by a different model. First, the intermediate response $\hat{z}_q$ is predicted using the model $f_1$ {(to be defined later)} as $\hat{z}_q = f_1(E_{ICL}^{(1)})_{d+1, D+1}$, where the input prompt is formatted as
\begin{equation}\label{eqn:input_icl}
E_{ICL}^{(1)}=\begin{pmatrix}
    x_1 & x_2 & \ldots & x_D & {x_{q}}\\
    z_1 & z_2 & \ldots & z_D & 0
\end{pmatrix}\in\mathbb{R}^{(d+1) \times (D+1)}.
\end{equation}
Obtaining $\hat{z}_q$, the input prompt of the second step is 
\begin{equation}\label{eqn:input_icl}
E_{ICL}^{(2)}=\begin{pmatrix}
    z_1 & z_2 & \ldots & z_D & \hat{z}_{q}\\
    y_1 & y_2 & \ldots & y_D & 0
\end{pmatrix}\in\mathbb{R}^{2 \times (D+1)}.
\end{equation}
The final prediction is obtained using the model $f_2$ as $\hat{y}_q = f_2(E_{ICL}^{(2)})_{d+1, D+1}$.
\end{definition}

\begin{definition}[Coherent CoT]\label{def:cot}
An Coherent CoT process involves two steps. In the first step, the input prompt is formatted as:
\begin{equation}\label{eqn:input_cot1}
E_{CoT}^{(1)}=\begin{pmatrix}
    x_1 & x_2 & \ldots & x_D & x_{q}\\
    z_1 & z_2 & \ldots & z_D & 0 \\
    y_1 & y_2 & \ldots & y_D & 0
\end{pmatrix}\in\mathbb{R}^{(d+2) \times (D+1)}.
\end{equation}
The intermediate response $\hat{z}_q$ is predicted as $\hat{z}_q = f(E_{CoT}^{(1)})_{d+1, D+1}$. After $\hat{z}_q$ is obtained, it is plugged into the input prompt, forming:
\begin{equation}
E_{CoT}^{(2)}=\begin{pmatrix}
    x_1 & x_2 & \ldots & x_D & x_{q}\\
    z_1 & z_2 & \ldots & z_D & \hat{z}_{q}\\
    y_1 & y_2 & \ldots & y_D & 0
\end{pmatrix}\in\mathbb{R}^{(d+2) \times (D+1)}.
\end{equation}
The final prediction of the query input is made by $\hat{y}_q = f(E_{CoT}^{(2)})_{d+2, D+1}$ {using the same transformer}.
\end{definition}

For both Stepwise ICL and Coherent CoT, we consider that the input prompt follows a structured data format. While some existing literature attempts to relax the assumption on the restrictive structured data format condition, e.g., \cite{xing2024benefits}, it is observed that
the performance of ICL with structured data serves as a lower bound for the best possible performance.

The main difference between Coherent CoT and Stepwise ICL is that, in the second step of Stepwise ICL, the input prompt only includes the intermediate and final responses, $z_i$s and $y_i$s, of the in-context examples. In contrast, Coherent CoT plugged the predicted $\hat{z}_q$ back into the original input prompt to form the input prompt for the second step and retains the initial inputs $x_i$s. 
This allows Coherent CoT to leverage both the intermediate reasoning and the original input when making the final prediction.

\paragraph{Model architecture.}
We follow \cite{zhang2023trained} and use transformers with one single-head linear attention layer as the model, which is defined as
\begin{equation}
    f(E)= W_{out}W^VE\cdot \left((W^KE)^{\top}W^QE\right),
\end{equation}
where $E$ denotes the input prompt, $W^Q$, $W^K$, $W^V \in \mathbb{R}^{m\times m}$ refer to the key, query and value matrix of the attention node and $W_{out} \in \mathbb{R}^{m\times m}$ refers to a fully-connected layer conducted to the output of the attention node. {For $f_1(\cdot)$ in Stepwise ICL, $m= d+1$; for $f_2(\cdot)$, $m = 2$. For CoT, $m=d+2$.}

\paragraph{Training objectives.} 
For both Stepwise ICL and Coherent CoT, to train a model, we minimize the following loss function:
\begin{eqnarray}\label{eqn:loss}
    L(\Theta)=\mathbb{E}_{\{x_i\},x_q}(\widehat y_q-y_q)^2,
\end{eqnarray}
which represents the mean squared error (MSE) between the predicted response $\widehat{y}_q$ and the true response $y_q$ of the query example. Here, $\Theta$ denotes the set of parameters {in the transformer}.

\subsection{Coherent CoT vs Stepwise ICL} \label{sec:theory}
This section presents the main results of comparing Coherent CoT and Stepwise ICL. To simplify the derivation, we assume a specific format for the optimal attention parameters as follows:

\begin{assumption}\label{ass:opt}
   We considerthe following specific formulations of the matrices $W^K$, $W^Q$, $W_{out}$ and $W^V$ for Stepwise ICL and Coherent CoT. 
   \begin{itemize}
    \vspace{-0.1in}
       \item For Stepwise ICL, the specific format for the parameters of $f_1(\cdot)$ is: $(W^{K})^\top W^Q=\begin{bmatrix}
    u_xI_d  &0 \\
     0 &  0  \\
\end{bmatrix}$ and  $(W_{out}W^V)_{d+1,:}=(0,\ldots,0, 1/u_x)$. For $f_2(\cdot)$, we assume $(W^{K})^\top W^Q=\begin{bmatrix}
    u_z  &0 \\
     0 &  0  \\
\end{bmatrix}$ and  $(W_{out}W^V)_{2,:}=(0, 1/u_z)$.
\item For Coherent CoT, we assume $$(W^{K})^\top W^Q=\begin{bmatrix}
    v_xI_d &  &0 \\
     & v_z &  \\
     0 & &v_y  \\
\end{bmatrix},$$ $$(W_{out}W^V)_{d+1,:}=(0,\ldots,0,1/v_x,0)$$
$$(W_{out}W^V)_{d+2,:}=(0,\ldots,0, 1/v_y).$$ 
   \end{itemize}
\end{assumption}

Assumption~\ref{ass:opt} is built upon the observations in~\cite{zhang2023trained,cui2024superiority}.
Based on these studies, in order to optimize the task of $x \rightarrow z$ in ICL, the corresponding off-diagonal elements of $(W^K)^\top W^Q$ and the first $d$ elements in the $(W_{out}W^V)_{d+1,:}$ vector should be all zero when $x$ follows $N(0,I_d)$. {Since each step of the Coherent CoT process utilizes the same transformer model, we consider the same format for the $z \rightarrow y$ process, and assume all off-diagonal elements of $(W^K)^\top W^Q$ are zero and $(W_{out}W^V)_{d+1,:}=(0,\ldots,0,v_y)$.}
With Assumption \ref{ass:opt}, we can rewrite $L(\Theta)$ to $L(u_x,u_z)$ for Stepwise ICL or $L(v_x,v_y,v_z)$ for Coherent CoT to highlight these parameters. 

Given Assumption~\ref{ass:opt}, we figure out the optimal solution of Stepwise ICL in Theorem~\ref{thm:icl}.

\begin{theorem}\label{thm:icl}
Under Assumption~\ref{ass:gen} and~\ref{ass:opt}, the optimal expected loss of Stepwise ICL is achieved when $u_x \neq 0$ and $u_z \neq 0$ are satisfied. The corresponding  optimal loss is 
\[L(u_x^*, u_z^*)= \sigma^2 + \frac{7+d}{D} + \frac{\sigma^2}{D} + o\left(\frac{1}{D}\right).\]
\end{theorem}

The proof of Theorem~\ref{thm:icl}
can be found in Appendix~\ref{proof:icl}. In short, we separate the MSE loss of the prediction into multiple terms and taking the expectation of each term considering $D \rightarrow \infty$.

\begin{remark}\label{rem:gaussian}
In Assumption \ref{ass:gen}, we assume that $x\sim N(0,I_d)$. The exact Gaussian distribution is used to derive the closed-form expression of the loss. If we relax it to other distributions, there will be no closed-form expression of the loss to exactly compare Coherent CoT and Stepwise ICL. Nonetheless, the high-level intuition on why Coherent CoT outperforms Stepwise ICL still holds: Stepwise ICL misses information on the previous reasoning steps, thus the prediction is worse than Coherent CoT.  
\end{remark}

To compare with the optimal results of Stepwise ICL, we derive the optimal solution for {Coherent CoT} and present the results in Theorem~\ref{thm:cot}.
\begin{theorem}\label{thm:cot}
    Under Assumption~\ref{ass:gen} and~\ref{ass:opt}, the optimal parameters $v_x$, $v_y$ and $v_z$ that minimize the Coherent CoT's loss satisfying $v_y=\left(v_x+v_z\right)$. The corresponding loss for Coherent CoT becomes
    \begin{eqnarray*}
      L(v_x,v_y,v_z)&=&\sigma^2 + \frac{d\sigma^2 }{D} + \frac{1+d}{D} +\frac{4}{D}  \frac{v_xv_z}{v_y^2}  + \frac{6}{D}\frac{v_z^2}{v_y^2}  \\
      &&   - \frac{v_z(2v_x+v_z)}{D v_y^2} (d-1)\sigma^2 +o\left(\frac{1}{D}\right). 
    \end{eqnarray*}

    To minimize the above loss, it is require that $v_y=\frac{(d-1)\sigma^2+2}{(d-1)\sigma^2-2}v_z = \frac{(d-1)\sigma^2+2}{4}v_x$ (where $v_z, v_x, v_y\neq0$). Then the optimal expected loss of Coherent CoT is 
\begin{eqnarray*}
L(v_x^*,v_y^*,v_z^*)=\sigma^2 + \frac{d\sigma^2 }{D} + \frac{1+d}{D}-\frac{\left((d-1)\sigma^2-2\right)^2}{D((d-1)\sigma^2+2) }+o\left(\frac{1}{D}\right). 
        \end{eqnarray*}
        \vspace{-0.25in}
\end{theorem}
The proof of Theorem~\ref{thm:cot}
is similar to that of Theorem~\ref{thm:icl}, whose details are shown in Appendix~\ref{proof:cot}. 

Building on the above results, Proposition \ref{prop:comp} below directly compares the expected loss of Coherent CoT and Stepwise ICL.

\begin{proposition}\label{prop:comp}
    Given that $d\geq 2$, the expected loss of Coherent CoT equals to the expected loss of Stepwise ICL when $v_x = 0$ and $v_z = v_y$. {In addition, the minimal expected loss of Coherent CoT is smaller than the one of Stepwise ICL.}
\end{proposition}
The proof of Proposition \ref{prop:comp} is in Appendix~\ref{proof:comp}. Proposition~\ref{prop:comp} indicates that, regardless of the values of $d$ and $\sigma$, the optimal expected loss of Coherent CoT is smaller than that of Stepwise ICL. To explain this, when $v_x = 0$ and $v_z = v_y$, Coherent CoT is equivalent to Stepwise ICL, and this condition also aligns with the optimal solution of Stepwise ICL. On the other hand, since $(v_x = 0,v_z = v_y)$ is not the optimal solution of Coherent CoT, Coherent CoT can achieve a smaller loss given its corresponding optimal solution.

\textbf{Insights from the theory.} To further investigate how the $x_i$s and $z_i$s contribute to the final prediction of $\hat{y}_q$ in CoT, we present the following proposition:

\begin{proposition}\label{prop:vs}
For any  $\sigma$ and $d$, $v_z$ and $v_y$ always share the same sign for optimal Coherent CoT. In addition, the ratio $\frac{v_z^*}{v_y^*}$ for the optimal Coherent CoT is consistently smaller $1$. Meanwhile, when $v_x$, $v_z$ and $v_y$ are set such that Coherent CoT reduces to Stepwise ICL, the ratio satisfies $\frac{v_z}{v_y}=1$.
\end{proposition}

{Based on Theorem \ref{thm:cot} and Proposition \ref{prop:vs}, the prediction of the final response can be formulated as }
\begin{eqnarray}
 \hat{y}_q = \frac{1}{D} \sum y_i\left(\frac{v_x}{v_y} x_i^\top x_q+\frac{v_z}{v_y}z_i{\hat{z_q}}\right).  
\end{eqnarray}The above formulation indicates how $x_i$s and $y_i$s impact the final prediction of $\hat{y}_q$.
According to Proposition \ref{prop:vs}, in Coherent CoT, the initial inputs $x_i$s always positively contribute to the final prediction $\hat{y}_q$. Additionally, the final prediction $\hat{y}_q$ in Coherent CoT relies less on the $z_i$s and $\hat z_q$ compared to Stepwise ICL, indicating a reduced dependency on these intermediate values when using the optimal Coherent CoT. 

Based on Proposition \ref{prop:vs}, the consequence of leveraging Coherent CoT to train a model is that, when there is an error in the prediction of $z_q$, Coherent CoT's reduced reliance on $\hat{z}_q$, along with its attention to  $x_q$, ensures that the error in the prediction of $z_q$ has a smaller impact on the final prediction. Moreover, since the final prediction of $y_q$ incorporates both $x_i$s and $z_i$s, when $\hat{z}_q$ is inaccurate, the model can better adjust its prediction using the $x_i$s values. This inherently provides a form of self-correction by leveraging the combined information from both $x_i$s and $z_i$s values.

Notably, the theorems in Section~\ref{sec:theory} and the later Section~\ref{sec:sensitivity} focus on different stages of the model's application. The theorem in Section~\ref{sec:theory} highlights that compared with Stepwise ICL, using Coherent CoT during the \textit{training stage} provides a better inference performance of the model. The theorem in Section~\ref{sec:sensitivity}, which examines the sensitivity of Coherent CoT to random noise, focuses on the \textit{inference stage}.    
  
\begin{figure}
    \centering
\includegraphics[width=0.82\linewidth]{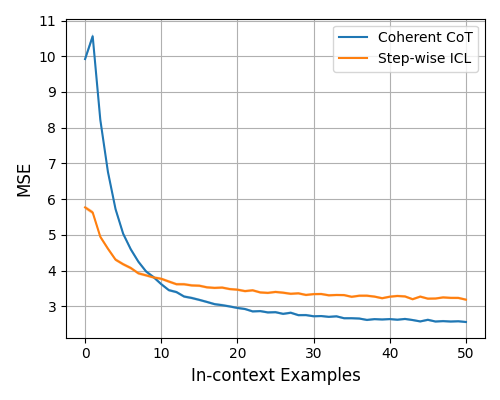}\vspace{-0.15in}
    \caption{\small Comparison of Coherent CoT and Stepwise ICL}
    \label{fig:simulation}
    \vspace{-0.1in}
\end{figure}

\textbf{Simulation Results.} {We also conduct simulations to demonstrate the advantage of Coherent CoT compared with Stepwise ICL. The transformers are trained from scratch to learn from the data generated from Assumption \ref{ass:gen}.} We modify the implementation of~\cite{li2023dissecting} to conduct the experiments. {To ensure stable convergence in training the Stepwise ICL, we separately trained the two models involved in the Stepwise ICL process. {Based on Theorem \ref{thm:icl}, when minimizing Eq.~\ref{eqn:loss}, Stepwise ICL indeed optimizes both transformers}}\footnote{More details of the setting of the simulation experiments can be found in Appendix~\ref{appd:sim}.}.
The results of the comparison are shown in Figure~\ref{fig:simulation}. {From the figure, we can observe that as the number of in-context examples increases, Coherent CoT converges to a lower error than Stepwise ICL. These results align with our theoretical findings, confirming the superiority of Coherent CoT in achieving more accurate predictions.}

\subsection{Sensitivity against Random Perturbation}\label{sec:sensitivity}
While Section~\ref{sec:theory} investigates how Coherent CoT gains a better prediction performance compared to Stepwise ICL by considering all the previous steps during the reasoning process, this holistic process may also lead to a different sensitivity to potential errors/corruptions in each reasoning step. 
Therefore, in Section \ref{sec:sensitivity}, {with a model trained with Coherent CoT,} we investigate the model's sensitivity and quantify the impact of random perturbations at different reasoning steps $y_i$, $x_i$, and $z_i$ at the inference stage. The results are summarized in the following theorems {respectively}.

  \begin{theorem}\label{thm:perturb_y}
   Under Assumption~\ref{ass:gen} and~\ref{ass:opt}, when there is random perturbation $\delta_i \sim N (0, \sigma_{\epsilon}^2)$ added to $y_i$, we denote the loss for Coherent CoT as $L'_{\text{y}}(v_x,v_y,v_z)$. When $v_x$, $v_y$ and $v_z$ takes the optimal values, we have \begin{eqnarray*}
       L'_{\text{y}}(v_x^*,v_y^*,v_z^*)-L(v_x^*,v_y^*,v_z^*)= o(1/D).
   \end{eqnarray*}
   \vspace{-0.3in}
\end{theorem}
  The proof of Theorem~\ref{thm:perturb_y} is postponed to Appendix~\ref{proof:perturb_y}. As indicated by the theorem, when the number of in-context examples D increases, the additional loss resulting from adding noise to $y$ decreases at a rate of $o(\frac{1}{D})$.  This suggests that adding noise to $y$ leads to a negligible influence on CoT performance.

While the above theorem discusses the sensitivity of CoT to random perturbation added to $y_i$s, the following theorem explores the cases when the random noise is added to $x_i$s.
  
  \begin{theorem}
  \label{thm:perturb_x}
 Under Assumption~\ref{ass:gen} and~\ref{ass:opt}, when there is a random perturbation $\delta_i \sim N (0, \sigma_{\epsilon}^2)$ added to $x_i$ in inference,  the expected loss for CoT becomes
      \begin{eqnarray*}
        &&  L'_{\text{x}}(v_x,v_y,v_z) \\&=& \frac{1+d}{D} +\frac{4}{D}  \frac{v_xv_z}{v_y^2}  + \frac{6}{D} \frac{v_z^2}{v_y^2}+\frac{d}{D}\sigma_{\epsilon}^2 +\sigma^2 +\frac{d\sigma^2}{D} \\&& - \frac{v_z(2v_x+v_z)}{D v_y^2} (d-1)\sigma^2 +\frac{v_x^2}{v_y^2}\sigma^2\sigma^2_\epsilon +o\left(\frac{1}{D}\right). 
      \end{eqnarray*}
      \vspace{-0.2in}
  \end{theorem}
The theorem below is for the corruptions in $z_i$s.
\begin{theorem}
  \label{thm:perturb_z}
 Under Assumption~\ref{ass:gen} and~\ref{ass:opt}, when there is a random perturbation $\delta_i \sim N (0, \sigma_{\epsilon}^2)$ added to $z_i$ in inference,  the expected loss for CoT becomes
\begin{align*}
   & L'_{\text{z}}(v_x,v_y,v_z)\\
    =& \frac{1+d}{D} +\frac{4}{D}  \frac{v_xv_z}{v_y^2}  + \frac{6}{D} \frac{v_z^2}{v_y^2} 
   +  \frac{3+d}{D} \frac{v_z^2}{v_y^2}\sigma_{\epsilon}^2+\sigma^2 +\frac{d\sigma^2}{D} \\& - \frac{v_z(2v_x+v_z)}{D v_y^2} (d-1)\sigma^2 +\frac{v_z^2}{v_y^2}\sigma^2\sigma^2_\epsilon +o\left(\frac{1}{D}\right). 
  \end{align*}
  
  \end{theorem}
The proof of Theorem~\ref{thm:perturb_x} and \ref{thm:perturb_z} are similar to that of Theorem~\ref{thm:perturb_y}, whose details are shown in Appendix~\ref{proof:perturb_x} and \ref{proof:perturb_z}.

Given the results in Theorem~\ref{thm:perturb_x} and~\ref{thm:perturb_z}, 
the following proposition provides a clearer comparison:
\begin{proposition}
When $v_x$, $v_y$ and $v_z$ takes the optimal values, 
 {we have  (1) $L'_{\text{x}}(v_x^*,v_y^*,v_z^*) > L'_{\text{y}}(v_x^*,v_y^*,v_z^*)$ and $L'_{\text{z}}(v_x^*,v_y^*,v_z^*) > L'_{\text{y}}(v_x^*,v_y^*,v_z^*)$ for any values of $d$ and $\sigma$; (2)} $L'_{\text{x}}(v_x^*,v_y^*,v_z^*) > L'_{\text{z}}(v_x^*,v_y^*,v_z^*)$ for $\sigma^2 \in \left[ 0, a\right) \cup({b,\infty}) $ and  $ L'_{\text{x}}(v_x^*,v_y^*,v_z^*) < L'_{\text{z}}(v_x^*,v_y^*,v_z^*)$ for $\sigma^2 \in \left( a, b \right)$, where $b > a$ are the positive roots of $f(\theta) = \left(d-1\right)^2 \theta^3+\left(3d-7\right)\left(d-1\right)\theta^2-\left(4d+8d^2\right)\theta+12 = 0$.\label{prop:perturb}
\end{proposition}

The proof of Proposition~\ref{prop:perturb} is shown in Appendix~\ref{proof:prop-perturb}.

\textbf{Insights from the theory.} The results in Proposition~\ref{prop:perturb} indicate that when a certain level of random noise is introduced to CoT, whether adding noise to the $x_i$s or the $z_i$s leads to a higher expected loss depends on the scale of the variance $\sigma^2$. Compared the results with that of adding noise to $y_i$, adding noise to $x_i$ or $z_i$ leads to a greater influence on the performance of CoT. An important insight from this result is that, compared to the noise in the final label (mislabeling), errors occurring in the reasoning steps of CoT have a greater influence on the accuracy of the final prediction. {From this perspective,} CoT is more sensitive to mistakes made during the reasoning process than to the noise in the final response. {As a result,} when asking the model to account for potential errors during reasoning, it is crucial to {pay} more attention to error {in the intermediate reasoning steps} rather than label noise.

\begin{figure}
    \centering
\includegraphics[width=0.82\linewidth]{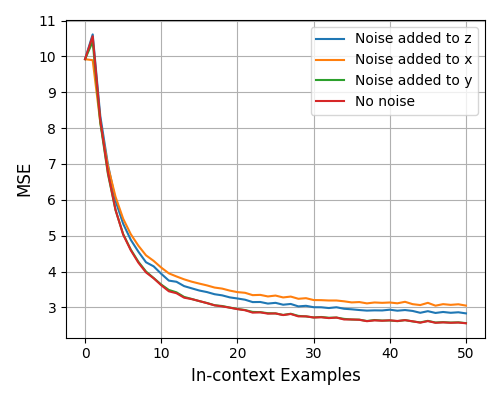}\vspace{-0.15in}
    \caption{\small CoT's sensitivity to different steps (The green and red lines overlap because the results for adding noise to $y$ and the case of no noise are very close to each other.)}
    \label{fig:simulation2}
    \vspace{-0.15in}
\end{figure}

\textbf{Simulation Results.} The simulation illustrating the sensitivity of CoT to random perturbations at different reasoning steps is shown in Figure~\ref{fig:simulation2}\footnote{In both Figure~\ref{fig:simulation} and~\ref{fig:simulation2} we plot the lower and upper bounds of the 90\% prediction interval for the mean loss values. However, when the bounds are very close to the mean, they may not be clearly visible.}. The training setup is similar to the experiment in Figure~\ref{fig:simulation}, with the variance of the noise, $\sigma^2_\epsilon$, set to 1.
From the figure, we observe that when random perturbations are added to the $y_i$s, the loss remains close to the case where there is no noise. However, when noise is introduced to the  $x_i$s or $z_i$s, the loss increases significantly. This result is consistent with our theoretical findings, confirming that CoT is more sensitive to perturbations in the earlier reasoning steps than label noise. 

\section{Improving CoT through Error-Aware Demonstrations}\label{sec:method}

The analysis in Section \ref{sec:sensitivity} reveals that, at the inference stage, CoT is more sensitive to errors in the intermediate reasoning steps of the demonstration than to errors in the final outcome. Inspired by this, we conjecture that
it is beneficial for the model to learn to adjust its predictions by considering the possibility of earlier mistakes. Therefore, we propose to incorporate both correct and incorrect reasoning paths {when composing} CoT demonstrations. 

\subsection{Methodology}\label{sec:proposed_method}

\begin{figure*}
    \centering 
    \vspace{-0.1in}
\includegraphics[width=0.93\linewidth]{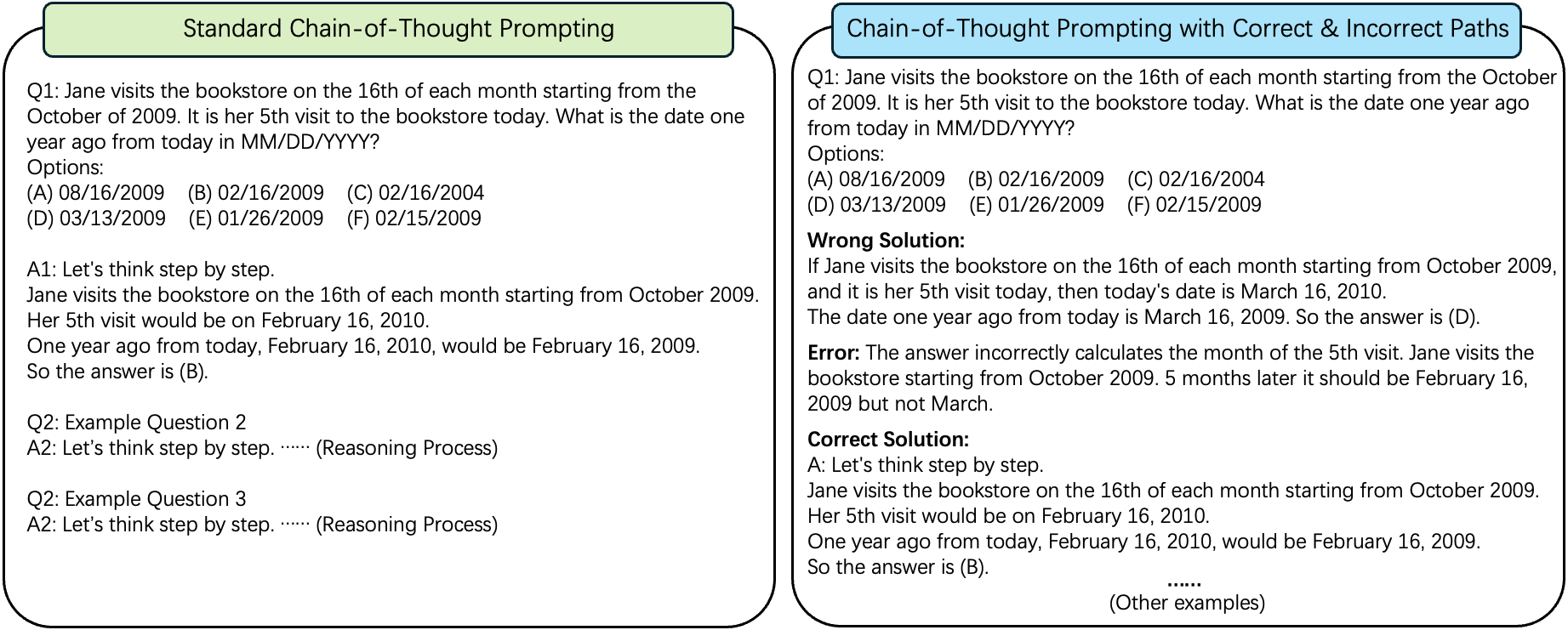}\vspace{-0.1in}
    \caption{Standard CoT prompting vs the proposed method: CoT prompting with correct \& incorrect paths. }
    \label{fig:flow}
    \vspace{-0.1in}
\end{figure*}

{We use a date understanding problem as an example to explain the proposed method in detail, and present the demonstration format in Figure~\ref{fig:flow}.}
In standard CoT prompting, few-shot examples with correct reasoning paths are provided in the prompt demonstration. In contrast, in the proposed demonstration format, after presenting the question, we first provide a potentially incorrect reasoning path, clearly labeled as a wrong solution. We then provide a detailed explanation of why this reasoning is incorrect, pointing out specific missteps or logical errors. Both the incorrect reasoning path and the analysis of why it is flawed are created manually. After analyzing the incorrect path, we provide a step-by-step correct reasoning process that leads to the correct answer.

By incorporating both correct and incorrect reasoning paths in the demonstrations, 
the proposed method teaches the model not only the correct reasoning path but also how to identify and handle potential reasoning errors.
This error-aware approach helps improve the model’s ability to adjust predictions and enhances its overall performance in reasoning tasks.

\subsection{Experiments}\label{sec:exp}
In this section, we evaluate our proposed method with various LLMs on multiple benchmarks.
\subsubsection{Experimental setup}\label{sec:exp_set}
\paragraph{Language Models.} Our experiments involve four LLMs: GPT-3.5-Turbo~\cite{brown2020language}, GPT-4o-mini~\cite{brown2020language}\footnote{The two GPT models are used under OpenAI's license.}, Gemini Pro~\cite{team2023gemini}\footnote{The model is used under Google's licensing terms.} and DeepSeek 67B~\cite{bi2024deepseek}\footnote{The model is used under DeepSeek's licensing terms.}. For generation, we set the temperature to 0 to ensure deterministic outputs.

\paragraph{Benchmarks.}

We use five datasets from two benchmarks for the experiments: the BBH benchmark~\cite{srivastava2022beyond} and the GSM8k benchmark~\cite{cobbe2021training}\footnote{Both datasets are distributed under the MIT license.}. The BBH benchmark focuses on reasoning tasks, and we select Disambiguation QA, Tracking Shuffled Objects (7 objects), Date Understanding, and Penguins in a Table from this benchmark. Each of these datasets contains 250 examples. Besides the BBH benchmark, we also use the GSM8k benchmark, which consists of 1,319 examples of arithmetic and grade-school math problems. 

\subsection{Main Results}
In Table \ref{tab:1}, we demonstrate the performance of various LLMs across different datasets when using standard CoT prompting (w/o IR) and our proposed method, which incorporates Incorrect Reasoning (IR) in the demonstrations (w/ IR).
From Table \ref{tab:1}, we can see that in most cases, adding handcrafted incorrect reasoning paths to CoT demonstrations improves the models' performance.
In some settings, our proposed method brings a significant improvement exceeding 5\%.
For example, in the Tracking Shuffled Objects dataset, Gemini Pro shows a 6.60\% improvement (from 58.20\% to 64.80\%), and in Penguins in a Table, DeepSeek 67B demonstrates an increase of 6.17\% (from 73.97\% to 80.14\%).
These results highlight the positive impact of exposing models to incorrect reasoning paths.

\begin{table}[h]
 \vspace{-0.1in}
\caption{\small Performance of CoT with and without handcrafted Incorrect Reasoning (IR) across different datasets and models.}\label{tab:1}
\vspace{0.05in}
\centering
 \resizebox{0.48\textwidth}{!}{
\begin{tabular}{cccc}
\toprule
\textbf{Dataset} & \textbf{Method} & \textbf{GPT-3.5-Turbo} & \textbf{GPT-4o-mini}  \\
\midrule
\multirow{2}{*}{\makecell{Disambiguation \\ QA}} & w/o IR & 68.00\% & 70.00\% \\ 
 & w/ IR &\textbf{ 72.00\%} & \textbf{71.60\%} \\ 
 \midrule
\multirow{2}{*}{\makecell{Tracking \\ Shuffled Objects}} & w/o IR & 56.53\% & 88.00\% \\ 
 & w/ IR & \textbf{61.20\%} & \textbf{88.80\%} \\ \midrule
\multirow{2}{*}{\makecell{Date\\ understanding}} & w/o IR & 82.27\% & 90.80\% \\ 
 & w/ IR & \textbf{85.07\%} &\textbf{ 91.47\% }\\ \midrule
\multirow{2}{*}{\makecell{Penguins \\ in a table}} & w/o IR & 81.34\% & 91.10\% \\ 
 & w/ IR & \textbf{82.19\%}&\textbf{92.92\% }  \\ \midrule
\multirow{2}{*}{GSM8K} & w/o IR & 81.03\% & 92.72\%  \\
 & w/ IR & \textbf{83.38}\% & \textbf{93.03\%}  \\ \bottomrule 
 \toprule
 \textbf{Dataset} & \textbf{Method} &  \textbf{Gemini Pro} & \textbf{DeepSeek 67B} \\
\midrule
\multirow{2}{*}{\makecell{Disambiguation \\ QA}} & w/o IR  & 68.80\% & 81.20\% \\ 
 & w/ IR & \textbf{76.80\%} & 81.20\% \\ 
 \midrule
\multirow{2}{*}{\makecell{Tracking \\ Shuffled Objects}} & w/o IR  & 58.20\% & 71.20\% \\ 
 & w/ IR  & \textbf{64.80\%} & \textbf{72.40\%} \\ \midrule
\multirow{2}{*}{\makecell{Date\\ understanding}} & w/o IR  & 88.80\% & {80.80\%} \\ 
 & w/ IR &  88.80\% & \textbf{83.20\%} \\ \midrule
\multirow{2}{*}{\makecell{Penguins \\ in a table}} & w/o IR  & 82.19\% & 73.97\% \\ 
 & w/ IR &\textbf{ 83.56\% }& \textbf{80.14\%} \\ \midrule
\multirow{2}{*}{GSM8K} & w/o IR & 80.82\% & \textbf{83.85\%} \\
 & w/ IR & \textbf{81.27\% }& 83.40\% \\ \bottomrule
\end{tabular}}
\vspace{-0.1in}
\end{table}

\subsection{Additional Experiments}

\begin{table}[h]
\vspace{-0.2in}
\caption{\small Performance of CoT with GPT-3.5-Turbo with and without model error explanation (EE).}\label{tab:abl}
\centering
 \resizebox{0.485\textwidth}{!}{
\begin{tabular}{cccccc} 
\toprule
{Method} &{\makecell{Disambig- \\ uation QA} } & {\makecell{Tracking Shu-\\ ffled Objects}} & {\makecell{Date under-\\ standing}} & {\makecell{Penguins \\ in a table}}  & {GSM8K}\\
\midrule
 w/o IR & 68.00\% & 56.53\% & 82.27\% & 81.34\% & 81.03\% \\ 
 \midrule
 w/ EE  & {72.00\%} & {61.20\%} & {85.07\%} & {82.19\%} & 83.38\% \\ 
  w/o EE & 70.80\% & {57.60\%} & {82.00\%} & {80.82\%} & {82.87\%} \\ 
  \bottomrule
\end{tabular}}
\vspace{-0.1in}
\end{table}

\textbf{Necessity of including error explanation.}
We present an ablation study to evaluate our proposed method when only the incorrect reasoning path is provided, without including an explanation of the error in that reasoning path. The experiments were conducted using GPT-3.5-Turbo, and the results on different datasets are shown in Table~\ref{tab:abl}. We compare the results with the case where both the incorrect reasoning paths and error explanations are included, as well as the case where neither of them is provided.

The results demonstrate that, in general, the performance of our method is reduced if the demonstration lacks an explanation of the error in the incorrect reasoning path. In some cases, such as with the Date Understanding dataset, the performance is even worse than when neither incorrect reasoning paths nor explanations are provided (from 82.27\% to 82.2\%). This suggests that merely presenting incorrect reasoning without explaining why it is wrong may confuse the model, making it harder for it to discern the correct logic from the incorrect one.
These results highlight the necessity of including error explanations when providing incorrect reasoning paths in demonstrations.

\textbf{Model-generated incorrect reasoning paths.}
As a variant of the proposed method, we further consider an implementation when the incorrect reasoning paths provided in the demonstration are not handcrafted but model-generated.
Specifically, we select incorrect reasoning paths from LLM-generated answers and include them in the demonstration. Intuitively, while handcrafting the wrong reasoning paths is generally more efficient, using model-generated incorrect paths has greater potential for improvement. Since the incorrect solutions are generated by the model itself, it allows the model to better recognize and avoid similar mistakes in future tasks. Notably, in this variant, the corresponding explanations of the errors in the reasoning still need to be provided manually.

The results when using model-generated incorrect reasoning paths are shown in Table~\ref{tab:2} as ``IR(M)", with a comparison to the handcrafted incorrect reasoning paths, denoted as ``IR (H)". The experiments are conducted on the Penguins in a Table Dataset.

From the results in Table~\ref{tab:2} we can observe that, across most models, using model-generated incorrect reasoning paths consistently leads to better performance compared to using handcrafted incorrect reasoning paths. In some cases, the improvement is substantial: for DeepSeek 67B, the improvement is from 82.88\% to 88.36\%. This indicates that model-generated incorrect paths allow the models to better recognize and learn from their own errors, resulting in enhanced reasoning capabilities. 

\begin{remark}
    The results for the w/o IR case in Table~\ref{tab:2} are different from those in Table~\ref{tab:1}. This is because the corresponding example questions used in the demonstrations are different.  This discrepancy suggests that, beyond incorporating incorrect reasoning paths, selecting appropriate example questions in the demonstration of CoT is also crucial.
\end{remark}

\begin{table}[h]
\centering
 \vspace{-0.15in}
\caption{\small Performance of CoT in the Penguins in a Table Dataset with or without handcrafted/model generated Incorrect Reasoning (IR(H)/IR(M)) across different models.}\label{tab:2}
 \resizebox{0.48\textwidth}{!}{
\begin{tabular}{ccccc}
\toprule
 {Method} & {\makecell{GPT-3.5- \\turbo}}& {\makecell{GPT-4o-\\mini} }& {\makecell{Gemini\\ Pro}} & {\makecell{DeepSeek \\67B} }\\
\midrule
 w/o IR  & 79.45\% & {99.32\%} & 81.51\% & 81.51\% \\ 
  w/ IR(H) & 87.67\% &  {99.32\%}  & 82.88\% & 82.88\% \\
 w/ IR(M) & \textbf{89.04\%} & {99.32\%} & \textbf{85.62\%} & \textbf{88.36\%} \\ 
  \bottomrule
\end{tabular}}
\vspace{-0.1in}
\end{table}

\section{Conclusion}
This paper provides a theoretical analysis of Coherent CoT by investigating its advantages over Stepwise ICL, showing that treating CoT as a holistic process leads to improved error correction and prediction accuracy.  In addition, we examine Coherent CoT's sensitivity to errors in different reasoning steps of the demonstration examples during the inference stage.
We observe that Coherent CoT is more sensitive to the error in the intermediate reasoning process than the inaccuracies in the final response. Inspired by this result, we propose to incorporate both correct and incorrect reasoning paths in demonstrations to improve the accuracy of the intermediate steps to enhance CoT performance. Experimental results validate the effectiveness of this approach.



\bibliographystyle{plain}
\bibliography{reference}

\appendix
\onecolumn

\section{Proofs}
\subsection{Proof of Theorem~\ref{thm:icl}:}\label{proof:icl}

The proof begins by decomposing the mean square loss into distinct components involving attention scores. Each term's expectation is then calculated separately, are combined to represent the total expected loss. 

Recalling that for Step-Wise ICL, we first obtain $\hat{z}_q$ by 

\begin{eqnarray*}
    \hat{z}_q &=& f_1(E_{ICL}^{(1)})_{d+1, D+1} \\
    &=& (W_{out}W^V_{d+1,:})
    E\left(\frac{1}{D}E^{\top}(W^K)^{\top} W^Q \begin{bmatrix}
        x_q \\ 0
    \end{bmatrix} \right) \\
&=&(\frac{1}{Du_x})\begin{bmatrix}
        z_1,z_2,\ldots,z_D,0
    \end{bmatrix}\left( \begin{bmatrix}
        u_x x_1^\top x_q\\ u_x x_2^\top x_q\\\vdots\\u_x x_q^\top x_q
    \end{bmatrix}  \right)\\
    &=& \frac{1}{D} \sum z_i(x_i^\top x_q)\\
\end{eqnarray*}
Then $\hat{y}_q$ is obtained by 
\begin{eqnarray*}
    \hat{y}_q &=& f_2(E_{ICL}^{(2)})_{2, D+1} \\
    &=& (W_{out}W^V_{d+1,:})
    E\left(\frac{1}{D}E^{\top}(W^K)^{\top} W^Q \begin{bmatrix}
        \hat{z}_q \\ 0
    \end{bmatrix} \right) \\
&=&(\frac{1}{Du_z})\begin{bmatrix}
        y_1,y_2,\ldots,y_D,0
    \end{bmatrix}\left( \begin{bmatrix}
        u_z z_1^\top \hat{z}_q\\ u_z z_2^\top \hat{z}_q\\\vdots\\u_z \hat{z}_q^\top \hat{z}_q
    \end{bmatrix}  \right)\\
    &=& \frac{1}{D} \sum y_i(z_i^\top \hat{z}_q) = \frac{1}{D^2} \sum y_i z_i^\top \left(\sum z_i(x_i^\top x_q)\right)\\
\end{eqnarray*}

Since $z_i=\beta^\top x_i$ and $y_i=\beta^\top x_i+\epsilon_i$, where $\epsilon_i\sim N(0,\sigma^2)$.
Then the expectation of the MSE loss can be expressed as: 
\begin{eqnarray}
  L(u_x, u_z)&=& \mathbb{E}\left(y_q-\hat{y}_q\right)^2 \nonumber \\
   &=& \mathbb{E}_{\{x_i\}_{i\in[D]}}\left(\left(\beta^\top x_q+\epsilon_q\right) - \frac{1}{D^2} \left(\sum \left(\beta^\top x_i+\epsilon_i\right) (\beta^\top x_i )\right)\left(\sum \beta^\top x_i(x_i^\top x_q)\right)\right)^2 \label{l_icl}\\
    &=& \mathbb{E}_{\{x_i\}_{i\in[D]}}\left(\beta^\top x_q+\epsilon_q\right)^2 +  \underbrace{\mathbb{E}_{\{x_i\}_{i\in[D]}}\left(\frac{1}{D^2} \left(\sum \left(\beta^\top x_i+\epsilon_i\right) (\beta^\top x_i )\right)\left(\sum \beta^\top x_i(x_i^\top x_q)\right)\right)^2}_{A_1} \nonumber \\
    && - \underbrace{\mathbb{E}_{\{x_i\}_{i\in[D]}}\left(\frac{2}{D^2} \left(\beta^\top x_q+\epsilon_q\right) \sum \left(\beta^\top x_i+\epsilon_i\right) (\beta^\top x_i )\left(\sum \beta^\top x_i(x_i^\top x_q)\right)\right)}_{A_2} \nonumber 
\end{eqnarray}
\begin{eqnarray*}
    \mathbb{E}_{\{x_i\}_{i\in[D]}}\left(\beta^\top x_q+\epsilon_q\right)^2 = \mathbb{E}_{\{x_i\}_{i\in[D]}}\left(\beta^\top x_q x_q^\top \beta+\epsilon_q^2 + 2\beta^\top x_q \epsilon_q \right) = \|\beta\|^2 + \sigma^2
\end{eqnarray*}
\begin{eqnarray*}
A_1&=&  \frac{(D-1)(D-2)(D-3)}{D^3}\mathbb{E}\left(\left(\beta^\top x_1 x_1^\top \beta + \beta^\top x_1\epsilon_1\right)\left(\beta^\top x_2 x_2^\top \beta + \beta^\top x_2\epsilon_2\right)\left(\beta^\top x_3x_3^\top x_q \right)\left(\beta^\top x_4x_4^\top x_q \right)\right)\\
&&+ \frac{(D-1)(D-2)}{D^3}\mathbb{E}\left(\left(\beta^\top x_1 x_1^\top \beta + \beta^\top x_1\epsilon_1\right)\left(\beta^\top x_2 x_2^\top \beta + \beta^\top x_2\epsilon_2\right)\left(\beta^\top x_3x_3^\top x_q \right)^2\right)\\
&&+ \frac{(D-1)(D-2)}{D^3}\mathbb{E}\left(\left(\beta^\top x_1 x_1^\top \beta + \beta^\top x_1\epsilon_1\right)^2\left(\beta^\top x_3x_3^\top x_q \right)\left(\beta^\top x_4x_4^\top x_q \right)\right) + o\left(\frac{1}{D}\right)\\
&&+\frac{4(D-1)(D-2)}{D^3}\mathbb{E}\left(\left(\beta^\top x_1 x_1^\top \beta + \beta^\top x_1\epsilon_1\right)\left(\beta^\top x_2 x_2^\top \beta + \beta^\top x_2\epsilon_2\right)\left(\beta^\top x_1x_1^\top x_q \right)\left(\beta^\top x_3x_3^\top x_q \right)\right)\\
&=&  \frac{(D-1)(D-2)(D-3)}{D^3}\mathbb{E}\left(\left(\beta^\top x_1x_1^\top\beta \right)\left(\beta^\top x_2x_2^\top\beta \right)\left(\beta^\top x_3x_3^\top x_q \right)\left(\beta^\top x_4x_4^\top x_q \right)\right)\\
&&+ \frac{(D-1)(D-2)}{D^3}\mathbb{E}\left(\left(\beta^\top x_1 x_1^\top \beta \right)\left(\beta^\top x_2 x_2^\top \beta \right)\left(\beta^\top x_3x_3^\top x_q  \right)^2\right) \\
&& + \frac{(D-1)(D-2)}{D^3}\mathbb{E}\left(\left((\beta^\top x_1 x_1^\top \beta)^2 + \beta^\top x_1x_1^\top \beta\epsilon_1^2\right)\left(\beta^\top x_3x_3^\top x_q \right)\left(\beta^\top x_4x_4^\top x_q  \right)\right) + o\left(\frac{1}{D}\right)\\
&&+\frac{4(D-1)(D-2)}{D^3}\mathbb{E}\left(\left(\beta^\top x_1 x_1^\top \beta\right)\left(\beta^\top x_2 x_2^\top \beta\right)\left(\beta^\top x_1x_1^\top x_q  \right)\left(\beta^\top x_3x_3^\top x_q  \right)\right)\\
&=& \frac{(D-1)(D-2)(D-3)}{D^3} \|\beta\|^6+\frac{(D-1)(D-2)}{D^3}(2+d) \|\beta\|^6 +\frac{(D-1)(D-2)}{D^3}\sigma^2 \|\beta\|^4 \\
&&+ \frac{15(D-1)(D-2)}{D^3} \|\beta\|^6 +o\left(\frac{1}{D}\right) \\
&=& \|\beta\|^6 +\frac{11}{D}\|\beta\|^6 + \frac{d}{D} \|\beta\|^6 + \frac{1}{D} \sigma^2 \|\beta\|^4+o\left(\frac{1}{D}\right)
\end{eqnarray*} 
\begin{eqnarray*}
    A_2 &=& \frac{2}{D} \mathbb{E}\left(\beta^\top x_q \beta^\top x_1x_1^\top \beta \beta^\top x_1 x_1^\top x_q\right) +\frac{2D(D-1)}{D^2}\mathbb{E}\left( \beta^\top x_1 x_1^\top \beta \beta^\top x_2 x_2^\top \beta \right)\\
    &=& \frac{4}{D} \|\beta\|^4 + 2 \|\beta\|^4 
\end{eqnarray*}
Based on the results of $A_1$ and $A_2$, we have 
\begin{eqnarray*}
  L(u_x^*, u_z^*)&=& \|\beta\|^2 + \sigma^2 +  \|\beta\|^6 +\frac{11}{D}\|\beta\|^6 + \frac{d}{D} \|\beta\|^6 + \frac{1}{D} \sigma^2 \|\beta\|^4 - \left(\frac{4}{D} \|\beta\|^4 + 2 \|\beta\|^4 \right)+o\left(\frac{1}{D}\right).  \end{eqnarray*}

Since $\|\beta\|^2 = 1$, we have
\begin{eqnarray*}
  L(u_x^*, u_z^*)&=&  \sigma^2 +\frac{7+d}{D} +\frac{\sigma^2}{D}
+o\left(\frac{1}{D}\right).   \end{eqnarray*}

As the parameters $u_x$ and $u_z$ are canceled out after taking the expectation, the optimal values $u_x^*$ and $u_z^*$  can be any non-zero values.

\subsection{Proof of Theorem~\ref{thm:cot}.}\label{proof:cot}

The proof begins by decomposing the mean square loss into distinct components involving attention scores. Each term's expectation is then calculated separately, are combined to represent the total expected loss. To minimize this loss, optimal values for the parameters $v_x$, $v_z$ and $v_y$  are derived. These optimal parameters are subsequently substituted back into the expression to further refine the loss representation.

The prediction of the final response $y_q$ can be expressed as:
    \begin{eqnarray*}
\hat{y}_q &=& (W_{out}W^V_{d+1,:})
    E\left(\frac{1}{D}E^{\top}(W^K)^{\top} W^Q \begin{bmatrix}
        x_q \\ \hat{z}_q\\0
    \end{bmatrix} \right) \\
&=&(\frac{1}{Dv_y})\begin{bmatrix}
        y_1,y_2,\ldots,y_D,0
    \end{bmatrix}\left( \begin{bmatrix}
        v_x x_1^\top x_q+v_z z_1^\top \hat{z}_q\\ v_x x_2^\top x_q+v_z z_2^\top \hat{z}_q\\\vdots\\v_x x_q^\top x_q+v_z z_q^\top \hat{z}_q
    \end{bmatrix}  \right)\\
    &=& (\frac{1}{Dv_y}) \sum y_i(v_x x_i^\top x_q+v_z z_i^\top \hat{z}_q)\\
\end{eqnarray*}

where 
    \begin{eqnarray*}
\hat{z}_q &=& (W_{out}W^V_{d,:})
    E\left(\frac{1}{D}E^{\top}(W^K)^{\top} W^Q \begin{bmatrix}
        x_q \\ 0 \\0
    \end{bmatrix} \right) \\
&=&(\frac{1}{Dv_x})\begin{bmatrix}
        z_1,z_2,\ldots,z_D,0
    \end{bmatrix}\left( \begin{bmatrix}
        v_x x_1^\top x_q \\ v_x x_2^\top x_q\\\vdots\\v_x x_q^\top x_q 
    \end{bmatrix}  \right)\\
    &=& \frac{1}{D} \sum z_ix_i^\top x_q\\
\end{eqnarray*}

Therefore, the expectation of the MSE loss can be expressed as: 
\begin{eqnarray*}
    \mathbb{E}\left(y_q-\hat{y}_q\right)^2 &=& \mathbb{E}_{\{x_i\}_{i\in[D]}}\left(\left(\beta^\top x_q + \epsilon_q \right)-\frac{1}{D}\frac{v_x}{v_y}\sum (\beta^\top x_i+\epsilon_i) x_i^\top x_q-\frac{1}{D^2}\frac{v_z}{v_y} \left(\sum  (\beta^\top x_i+\epsilon_i)  (\beta^\top x_i)\right) \left( \sum \beta^\top x_i x_i^\top x_q \right) \right)^2 \\
    &=& \mathbb{E}_{\{x_i\}_{i\in[D]}}\left(\beta^\top x_q + \epsilon_q \right)^2 +\underbrace{\frac{1}{D^2}\frac{v_x^2}{v_y^2} \left(\sum (\beta^\top x_i+\epsilon_i) x_i^\top x_q\right)^2 }_{A_1} \\
    &&+ \underbrace{\mathbb{E}_{\{x_i\}_{i\in[D]}}\left(\frac{1}{D^2} \frac{v_z}{v_y}\left(\sum \left(\beta^\top x_i+\epsilon_i\right) (\beta^\top x_i )\right)\left(\sum \beta^\top x_i(x_i^\top x_q)\right)\right)^2}_{A_2} \\
    && - \underbrace{\mathbb{E}_{\{x_i\}_{i\in[D]}}\left(\frac{2}{D} \frac{v_x}{v_y}\left(\beta^\top x_q+\epsilon_q\right) \sum \left(\beta^\top x_i+\epsilon_i\right)x_i^\top x_q \right)}_{A_3} \\
    && - \underbrace{\mathbb{E}_{\{x_i\}_{i\in[D]}}\left(\frac{2}{D^2} \left(\beta^\top x_q+\epsilon_q\right) \left(\sum \left(\beta^\top x_i+\epsilon_i\right) (\beta^\top x_i )\right) \left(\sum \beta^\top x_i(x_i^\top x_q)\right)\right)}_{A_4} \\
    && + \underbrace{\mathbb{E}_{\{x_i\}_{i\in[D]}}\left(\frac{2}{D^3} \frac{v_xv_z}{v_y^2}\left(\sum \left(\beta^\top x_i+\epsilon_i\right) x_i^\top x_q \right) \left(\sum \left(\beta^\top x_i+\epsilon_i\right) (\beta^\top x_i )\right)\left(\sum \beta^\top x_i(x_i^\top x_q)\right)\right)}_{A_5} 
\end{eqnarray*}

\begin{eqnarray*}
    \mathbb{E}_{\{x_i\}_{i\in[D]}}\left(\beta^\top x_q + \epsilon_q \right)^2 = \|\beta\|^2 +\sigma^2 
\end{eqnarray*}

\begin{eqnarray*}
    A_1 &=& \frac{v_x^2}{v_y^2}\frac{1}{D} \mathbb{E}\left((\beta^\top x_1 x_1^\top x_q)^2+ \epsilon_i^2(x_i^\top x_q)^2\right)+\frac{v_x^2}{v_y^2} \mathbb{E}\left( \frac{D(D-1)}{D^2} \beta^\top x_1 x_1^\top x_q x_q^\top x_2 x_2^\top \beta \right)\\
    &=&  \frac{v_x^2}{v_y^2} \frac{(d+2)}{D} \|\beta\|^2 + \frac{v_x^2}{v_y^2} \frac{d}{D} \sigma^2 + \frac{v_x^2}{v_y^2}\frac{D(D-1)}{D^2} \|\beta\|^2= \frac{v_x^2}{v_y^2} \frac{d}{D} \sigma^2+\frac{v_x^2}{v_y^2}\frac{D(D+1+d)}{D^2} \|\beta\|^2
\end{eqnarray*}

\begin{eqnarray*}
    A_2= \frac{v_z^2}{v_y^2} \|\beta\|^6 +\frac{11}{D}\frac{v_z^2}{v_y^2}\|\beta\|^6 + \frac{d}{D}\frac{v_z^2}{v_y^2} \|\beta\|^6 + \frac{1}{D}\frac{v_z^2}{v_y^2} \sigma^2 \|\beta\|^4+o\left(\frac{1}{D}\right)
\end{eqnarray*}
\begin{eqnarray*}
    A_4 = \frac{4}{D}\frac{v_z}{v_y}\|\beta\|^4 + 2\frac{v_z}{v_y} \|\beta\|^4 
\end{eqnarray*}
\begin{eqnarray*}
A_3=\mathbb{E}\left(2 \frac{v_x}{v_y}\left(\beta^\top x_q\right)\beta^\top x_1x_1^\top x_q\right) =2 \frac{v_x}{v_y}\|\beta\|^2
\end{eqnarray*}
\begin{eqnarray*}
    A_5&=&\mathbb{E}_{\{x_i\}_{i\in[D]}}\left(\frac{2}{D^3} \frac{v_xv_z}{v_y^2}\left(\sum \left(\beta^\top x_i+\epsilon_i\right) x_i^\top x_q \right) \left(\sum \left(\beta^\top x_i+\epsilon_i\right) (\beta^\top x_i )\right)\left(\sum \beta^\top x_i(x_i^\top x_q)\right)\right)\\
    &=& \frac{2(D-1)(D-2)}{D^2} \frac{v_xv_z}{v_y^2}(\beta^\top x_1 x_1^\top x_q)(\beta^\top x_2x_2^\top\beta )(\beta^\top x_3 x_3^\top x_q)\\
    &&+ \frac{2(D-1)}{D^2} \frac{v_xv_z}{v_y^2}(\beta^\top x_1 +\epsilon_1)^2x_1^\top x_q\beta^\top x_1\beta^\top x_2x_2^\top x_q + \frac{2(D-1)}{D^2} \frac{v_xv_z}{v_y^2}(\beta^\top x_1x_1^\top x_q)^2\beta^\top x_2x_2^\top \beta \\
    &&+ \frac{2(D-1)}{D^2} \frac{v_xv_z}{v_y^2}(\beta^\top x_1x_1^\top x_q)(\beta^\top x_2x_2^\top \beta) (\beta^\top x_2x_2^\top x_q)+o\left(\frac{1}{D}\right)\\
    &=&\frac{2v_xv_z}{v_y^2}\|\beta\|^4-\frac{3}{D}\frac{2v_xv_z}{v_y^2}\|\beta\|^4 + \frac{2}{D}\frac{v_xv_z}{v_y^2} \sigma^2 \|\beta\|^2 +\frac{3}{D}\frac{2v_xv_z}{v_y^2}\|\beta\|^4+\frac{2d+4}{D}\frac{v_xv_z}{v_y^2}\|\beta\|^4+\frac{3}{D}\frac{2v_xv_z}{v_y^2}\|\beta\|^4+o\left(\frac{1}{D}\right)\\
    &=& \frac{2v_xv_z}{v_y^2}\|\beta\|^4 + \frac{2}{D}\frac{v_xv_z}{v_y^2} \sigma^2 \|\beta\|^2+\frac{2d+4}{D}\frac{v_xv_z}{v_y^2}\|\beta\|^4+\frac{3}{D}\frac{2v_xv_z}{v_y^2}\|\beta\|^4+o\left(\frac{1}{D}\right)
\end{eqnarray*}
Based on the results of $A_1$ to $A_5$, we have 
\begin{eqnarray*}
    \mathbb{E}\left(y_q-\hat{y}_q\right)^2 &=& \|\beta\|^2 +\sigma^2 +\frac{v_x^2}{v_y^2} \frac{d}{D} \sigma^2+\frac{v_x^2}{v_y^2}\frac{D(D+1+d)}{D^2} \|\beta\|^2 + \frac{v_z^2}{v_y^2} \|\beta\|^6 +\frac{11}{D}\frac{v_z^2}{v_y^2}\|\beta\|^6 + \frac{d}{D}\frac{v_z^2}{v_y^2} \|\beta\|^6 \\
    && + \frac{1}{D}\frac{v_z^2}{v_y^2} \sigma^2 \|\beta\|^4 - \frac{4}{D}\frac{v_z}{v_y}\|\beta\|^4 - 2\frac{v_z}{v_y} \|\beta\|^4  -2\frac{v_x}{v_y}\|\beta\|^2 + \frac{2v_xv_z}{v_y^2}\|\beta\|^4 + \frac{2}{D}\frac{v_xv_z}{v_y^2} \sigma^2 \|\beta\|^2\\
    &&+\frac{2d+4}{D}\frac{v_xv_z}{v_y^2}\|\beta\|^4+\frac{3}{D}\frac{2v_xv_z}{v_y^2}\|\beta\|^4 +o\left(\frac{1}{D}\right)\\
    &=& \sigma^2+ \mathbb{E}\left(\frac{\left(v_x-v_y+v_z\|\beta\|^2\right)^2\|\beta\|^2}{v_y^2} +\frac{v_x^2d+v_z^2\|\beta\|^4+2v_xv_z\|\beta\|^2}{D v_y^2}\sigma^2  \right)\\    &+&\mathbb{E}\left(\frac{1}{D}\frac{(v_x+v_z\|\beta\|^2)^2}{v_y^2} (1+d)\|\beta\|^2+\frac{10}{D}\frac{v_z^2}{v_y^2} \|\beta\|^6 +\frac{8}{D}\frac{v_xv_z}{v_y^2} \|\beta\|^4-\frac{4}{D}\frac{v_z}{v_y}
\|\beta\|^4\right)+o(\frac{1}{D})\\
\end{eqnarray*}

{ If we have $\|\beta\|^2 = 1$, it becomes


\begin{eqnarray*}
    \mathbb{E}\left(y_q-\hat{y}_q\right)^2
    &=&  \left(\frac{\left(v_x-v_y+v_z \right)^2}{v_y^2} + \sigma^2 +\frac{(v_x+v_z)^2 }{v_y^2} (\frac{1+d}{D}) -\frac{4}{D} \frac{v_zv_y}{v_y^2} +\frac{8}{D}  \frac{v_xv_z}{v_y^2}  +\frac{10}{D}  \frac{v_z^2}{v_y^2} +\frac{v_x^2d+v_z^2+2v_xv_z}{D v_y^2}\sigma^2 \right)+o(\frac{1}{D})\\
\end{eqnarray*}

To minimize the loss, we need to let $\left(v_x-v_y+v_z\right)^2 = 0$, meaning that $v_y =  (v_x+v_z)$

Then the loss becomes 
\begin{eqnarray*}
\mathbb{E}\left(y_q-\hat{y}_q\right)^2 &=&\sigma^2 + \frac{d\sigma^2 }{D} + \frac{1+d}{D} - \frac{v_z(2v_x+v_z)}{D v_y^2} (d-1)\sigma^2 +\frac{4}{D}  \frac{v_xv_z}{v_y^2}  + \frac{6}{D} \frac{v_z^2}{v_y^2} \\ &=&\sigma^2 + \frac{d\sigma^2 }{D} + \frac{1+d}{D} - \frac{(2v_y-v_z)v_z}{D v_y^2} (d-1)\sigma^2 +\frac{4}{D}  \frac{(v_y-v_z)v_z}{v_y^2}  + \frac{6}{D} \frac{v_z^2}{v_y^2} \\
\end{eqnarray*}

Letting $\theta$ represents $\frac{v_x}{v_y}$, the expectation can be further expressed as a function of $\theta$: 
\begin{eqnarray*}
\mathbb{E}\left(y_q-\hat{y}_q\right)^2 = f(\theta)&=&\sigma^2 + \frac{d\sigma^2 }{D} + \frac{1+d}{D} - \frac{(2\theta-\theta^2)}{D} (d-1)\sigma^2 +\frac{4}{D} (\theta-\theta^2)  + \frac{6}{D}\theta^2 \\
\end{eqnarray*}
Then we have 
\begin{eqnarray*}
f'(\theta)&=&  \frac{(2\theta-2)}{D} (d-1)\sigma^2 +\frac{1}{D} (4\theta+4)  \\
\end{eqnarray*}
Letting $f'(\theta)=0$, we have
\begin{eqnarray*}
  \theta = \frac{(d-1)\sigma^2-2}{(d-1)\sigma^2+2}
\end{eqnarray*}
Therefore, the optimal $\frac{v_z}{v_y}$ satisfies $\frac{v_z^*}{v_y^*} = \frac{(d-1)\sigma^2-2}{(d-1)\sigma^2+2}$. Then we further have $\frac{v_x^*}{v_y^*} = 1-\frac{v_z}{v_y}= \frac{4}{(d-1)\sigma^2+2}$.
Substituting the optimal parameters back to the formula of the expected loss, we have
\begin{eqnarray*}
    L(v_x^*,v_y^*,v_z^*) &=& \mathbb{E}\left(y_q-\hat{y}_q\right)^2=\sigma^2 + \frac{d\sigma^2 }{D} + \frac{1+d}{D}-\frac{\left((d-1)^3\sigma^6-2(d-1)^2\sigma^4-4(d-1)\sigma^2+12\right)}{((d-1)\sigma^2+2)^2}+o\left(\frac{1}{D}\right)\\
  &=&\sigma^2 + \frac{d\sigma^2 }{D} + \frac{1+d}{D}-\frac{\left((d-1)\sigma^2-2\right)^2}{D((d-1)\sigma^2+2) }+o\left(\frac{1}{D}\right).
\end{eqnarray*}
}

\subsection{Proof of Proposition~\ref{prop:comp}.}\label{proof:comp}
Recalling that the loss of Coherent CoT is written as:
\begin{eqnarray*}
   L(v_x,v_z,v_y) &=& \mathbb{E}_{\{x_i\}_{i\in[D]}}\left(\left(\beta^\top x_q + \epsilon_q \right)-\frac{1}{D}\frac{v_x}{v_y}\sum (\beta^\top x_i+\epsilon_i) x_i^\top x_q-\frac{1}{D^2}\frac{v_z}{v_y} \left(\sum  (\beta^\top x_i+\epsilon_i)  (\beta^\top x_i)\right) \left( \sum \beta^\top x_i x_i^\top x_q \right) \right)^2 \\
    \end{eqnarray*}

When setting $v_x = 0$, $v_y=v_z$, we have 
\begin{eqnarray*}
   L(v_x,v_z,v_y) &=& \mathbb{E}_{\{x_i\}_{i\in[D]}}\left(\left(\beta^\top x_q + \epsilon_q \right)-\frac{1}{D^2} \left(\sum  (\beta^\top x_i+\epsilon_i)  (\beta^\top x_i)\right) \left( \sum \beta^\top x_i x_i^\top x_q \right) \right)^2 \\
    \end{eqnarray*}
    which is equivalent to the Eq.(\ref{l_icl}) in the proof of Theorem~\ref{thm:icl}. This indicates that when setting $v_x = 0$, $v_y=v_z$, Coherent is equivalent to Step-wise CoT.
    Moreover, since $(v_x = 0,v_z = v_y)$ is not the optimal solution of Coherent CoT, Coherent CoT can achieve a smaller loss given its corresponding optimal solution. Therefore, we can conclude that the
minimal expected loss of Coherent CoT must be smaller than the one of Stepwise ICL.
    
\subsection{Proof of Theorem~\ref{thm:perturb_y}.}
The proof of Theorem~\ref{thm:perturb_y} to Theorem~\ref{thm:perturb_z} is similar to that of Theorem~\ref{thm:cot}, with the only difference being the presence of additional terms resulting from the added noise. To save space, we present only the additional terms introduced by the perturbation and denote the remaining terms as $L(v_x,v_y,v_z)$.

When there is random perturbation $\delta_i \sim N (0, \sigma_{\epsilon}^2)$ added to $y_i$, the loss becomes

 \begin{eqnarray*}
&& L'_{\text{y}}(v_x,v_y,v_z) = \mathbb{E}\left(y_q-\hat{y}_q\right)^2 \\
 &=& \mathbb{E}_{\{x_i\}_{i\in[D]}}\left(\left(\beta^\top x_q + \epsilon_q \right)-\frac{1}{D}\frac{v_x}{v_y}\sum (\beta^\top x_i+\epsilon_i+\delta_i) x_i^\top x_q-\frac{1}{D^2}\frac{v_z}{v_y} \left(\sum  (\beta^\top x_i+\epsilon_i+\delta_i)  (\beta^\top x_i)\right) \left( \sum \beta^\top x_i x_i^\top x_q \right) \right)^2 \\
 &=&  L(v_x,v_z,v_y) + \mathbb{E}\left( \frac{1}{D^2} \frac{v_x^2}{v_y^2} (\delta_ix_i^\top x_q)^2 + \frac{1}{D^4} \left(\sum \delta_i \beta^\top x_i \sum \beta^\top x_ix_i^\top x_q \right)^2 \right) \\
    &&+ \mathbb{E}\left(\frac{2}{D^3}\frac{v_xv_z}{v_y^2} \sum \delta_ix_i^\top x_q \sum \delta_i\beta^\top x_i \sum \beta^\top x_ix_i^\top x_q \right) + o\left(\frac{1}{D}\right) \\
   &=&   L(v_x,v_z,v_y) + \mathbb{E}\left(\frac{1}{D} \frac{v_x^2}{v_y^2} \delta_1^2 x_i^\top x_q x_q^\top x_i  + \frac{(D-1)(D-2)}{D^3} \frac{v_z^2}{v_y^2}\delta_1^2 \beta^\top x_1x_1^\top \beta \beta^\top x_2x_2^\top x_q\beta^\top x_3x_3^\top x_q \right)  \\
   &&+2 \mathbb{E}\left(\frac{(D-1)}{D^2} \frac{2v_xv_z}{v_y}\delta_1^2 x_1^\top x_q \beta^\top x_1 \beta^\top x_2x_2^\top x_q\right)+ o\left(\frac{1}{D}\right) \\
   &=&  L(v_x,v_z,v_y) + \frac{1}{D}\left(\frac{v_x-v_y+v_z}{v_y}\right)^2\sigma_\epsilon^2  + o\left(\frac{1}{D}\right)  \\
&=&  L(v_x,v_z,v_y) + o\left(\frac{1}{D}\right) 
    \end{eqnarray*}

\label{proof:perturb_y}
\subsection{Proof of Theorem~\ref{thm:perturb_x}.}\label{proof:perturb_x}
When there is random perturbation $\delta_i \sim N (0, \sigma_{\epsilon}^2)$ added to $x_i$, the loss becomes

\begin{eqnarray*}
   L'_{\text{x}}(v_x,v_y,v_z)= \mathbb{E}\left(y_q-\hat{y}_q\right)^2 &=& \mathbb{E}_{\{x_i\}_{i\in[D]}}\left(y_q-\frac{1}{D}\frac{v_x}{v_y}\sum y_i (x_i+\delta_i)^\top x_q-\frac{1}{D}\frac{v_z}{v_y}\sum y_i z_i \hat{z}_q\right)^2,\\
    \end{eqnarray*}
where
\begin{eqnarray*}
\hat{z}_q=\frac{1}{D}\sum z_i (x_i^\top+\delta_i^\top) x_q.
\end{eqnarray*}
Therefore, the loss can be written as: 
\begin{eqnarray*}
   &&  L'_{\text{x}}(v_x,v_y,v_z) = \mathbb{E}\left(y_q-\hat{y}_q\right)^2\\
    &=& \mathbb{E}_{\{x_i\}_{i\in[D]}}\left(\left(\beta^\top x_q + \epsilon_q \right)-\frac{1}{D}\frac{v_x}{v_y}\sum (\beta^\top x_i+\epsilon_i) (x_i^\top+\delta_i^\top) x_q-\frac{1}{D^2}\frac{v_z}{v_y} \left(\sum  (\beta^\top x_i+\epsilon_i) \beta^\top x_i\right) \left( \sum \beta^\top x_i (x_i^\top+\delta_i^\top) x_q \right) \right)^2 \\
    &=&L(v_x,v_z,v_y) +\mathbb{E}\left( \frac{1}{D}\frac{v_x}{v_y}\sum \beta^\top x_i\delta_i^\top x_q \right)^2+\mathbb{E}\left( \frac{1}{D}\frac{v_x}{v_y}\sum \epsilon_i \delta_i^\top x_q \right)^2 +\left(\frac{1}{D}\frac{v_z}{v_y}\sum \beta^\top x_iz_i^\top (\frac{1}{D}\sum \beta^\top x_i \delta_i^\top x_q)\right)^2 \\
    &&+2\mathbb{E}\left[ \left(\frac{1}{D^2}\frac{v_xv_z}{v_y}\sum \beta^\top x_i \delta_i^\top x_q \right) \left(\sum \beta^\top x_iz_i^\top (\frac{1}{D}\sum \beta^\top x_i \delta_i^\top x_q)\right)\right]^2 +o(\frac{1}{D}) \\
    &=& \left(\frac{1+d}{D} +\frac{4}{D}  \frac{v_xv_z}{v_y^2}  + \frac{6}{D} \frac{v_z^2}{v_y^2}+\sigma^2 +\frac{d\sigma^2}{D}  - \frac{v_z(2v_x+v_z)}{D v_y^2} (d-1)\sigma^2 \right)\\
&&+ \mathbb{E}\left(\frac{1}{D}\frac{v_x^2}{v_y^2} (\beta^\top x_1 \epsilon_1 ^\top x_q)^2+\frac{1}{D}\frac{v_x^2}{v_y^2} (\delta_1 \epsilon_1 ^\top x_q)^2 + \frac{2}{D}\frac{v_xv_z}{v_y^2} \beta^\top x_1 x_1^\top \beta (\beta^\top x_2 \epsilon_2 ^\top x_q)^2 \right) \\
    &&+\mathbb{E}\left(\frac{1}{D}\frac{v_z^2}{v_y^2} (\beta^\top x_1 x_1^\top \beta) (\beta^\top x_2 x_2^\top \beta) (\beta^\top x_3 \epsilon_3 ^\top x_q)^2 \right)+o(\frac{1}{D}) \\
    &=&  \frac{1+d}{D} +\frac{4}{D}  \frac{v_xv_z}{v_y^2}  + \frac{6}{D} \frac{v_z^2}{v_y^2}+\sigma^2 +\frac{d\sigma^2}{D}  - \frac{v_z(2v_x+v_z)}{D v_y^2} (d-1)\sigma^2 + \frac{d}{D}\frac{(v_x+v_z)^2}{v_y^2}\sigma_\epsilon^2+\frac{v_x^2}{v_y^2}\sigma^2\sigma^2_\epsilon+o(\frac{1}{D}) \\
    &=& \frac{1+d}{D} +\frac{4}{D}  \frac{v_xv_z}{v_y^2}  + \frac{6}{D} \frac{v_z^2}{v_y^2}+\frac{d}{D}\sigma_{\epsilon}^2 +\sigma^2 +\frac{d\sigma^2}{D}  - \frac{v_z(2v_x+v_z)}{D v_y^2} (d-1)\sigma^2 +\frac{v_x^2}{v_y^2}\sigma^2\sigma^2_\epsilon +o\left(\frac{1}{D}\right)\\
\end{eqnarray*}

\subsection{Proof of Theorem~\ref{thm:perturb_z}.}\label{proof:perturb_z}
When there is random perturbation $\delta_i \sim N (0, \sigma_{\epsilon}^2)$ added to $z_i$, the loss becomes

\begin{eqnarray*}
  L'_{\text{z}}(v_x,v_y,v_z)=  \mathbb{E}\left(y_q-\hat{y}_q\right)^2 &=& \mathbb{E}_{\{x_i\}_{i\in[D]}}\left(y_q-\frac{1}{D}\frac{v_x}{v_y}\sum y_i x_i^\top x_q-\frac{1}{D}\frac{v_z}{v_y}\sum y_i(z_i+\delta_i) \hat{z}_q\right)^2,\\
    \end{eqnarray*}

where
\begin{eqnarray*}
\hat{z}_q=\frac{1}{D}\sum (z_i +\delta_i)x_i^\top x_q.
\end{eqnarray*}
Therefore, the loss can be written as: 
\begin{eqnarray*}
    &&L'_{\text{z}}(v_x,v_y,v_z)=\mathbb{E}\left(y_q-\hat{y}_q\right)^2\\
    &=& \mathbb{E}_{\{x_i\}_{i\in[D]}}\left(\left(\beta^\top x_q + \epsilon_q \right)-\frac{1}{D}\frac{v_x}{v_y}\sum (\beta^\top x_i+\epsilon_i) x_i^\top x_q-\frac{1}{D^2}\frac{v_z}{v_y} \left(\sum  (\beta^\top x_i+\epsilon_i)  (\beta^\top x_i+\delta_i)\right) \left( \sum (\beta^\top x_i +\delta_i) x_i^\top x_q \right) \right)^2 \\
        &=& L(v_x, v_y, v_z)  -2y_q \left[\frac{1}{D^2}\frac{v_z}{v_y}\sum \left(\beta^\top x_i  \delta_i (\sum  \delta_i x_i^\top x_q ) \right)\right]+2  \left[\frac{1}{D^2}\frac{v_xv_z}{v_y^2} (\sum \beta^\top x_i x_i^\top x_q) \sum \left(\beta^\top x_i  \delta_i (\sum   \delta_i x_i^\top x_q ) \right)\right] \\
        && +\frac{1}{D^4} \frac{v_z^2}{v_y^2} \left( \left(\sum \epsilon_i \delta_i\right)^2  \left( \sum (\beta^\top x_i +\delta_i) x_i^\top x_q \right)^2 \right)\\
    &&+\frac{1}{D^2}\frac{v_z^2}{v_y^2}\left(\mathbb{E}\left[\sum \left(\beta^\top x_i (z_i+ \delta_i)^\top \sum (z_i + \delta_i)x_i^\top x_q\right) \right]^2-\mathbb{E}\left[\sum \left(\beta^\top x_i z_i \sum z_i x_i^\top x_q\right) \right]^2\right) +o(\frac{1}{D})\\
    &=& \left(\sigma^2 + \frac{d\sigma^2 }{D} + \frac{1+d}{D} - \frac{(2v_y-v_z)v_z}{D v_y^2} (d-1)\sigma^2 +\frac{4}{D}  \frac{v_xv_z}{v_y^2}  + \frac{6}{D} \frac{v_z^2}{v_y^2}\right) \\
    &&- \mathbb{E}\left(\frac{2}{D}\frac{v_zv_y}{v_y^2} \delta_1^2 \beta^\top x_q \beta^\top x_1 x_1^\top x_q+\frac{2}{D}\frac{v_xv_z}{v_y^2} \beta^\top x_1 x_1^\top x_q \beta^\top x_2 x_2^\top x_q \delta_2^2 \right) \\
    && + 
    \mathbb{E}\left( \frac{1}{D}\frac{v_z^2}{v_y^2} \epsilon_1^2 \delta_1^2 \beta^\top x_2x_2^\top x_q\beta^\top x_3x_3^\top x_q  + \frac{1}{D}\frac{v_z^2}{v_y^2} \beta^\top x_1x_1^\top\beta\delta_1^2 \beta^\top x_2x_2^\top x_q\beta^\top x_3x_3^\top x_q  \right) \\
    && + 
   \mathbb{E}\left( \frac{1}{D}\frac{v_z^2}{v_y^2} \beta^\top x_1x_1^\top \beta \beta^\top x_2x_2^\top \beta \delta_3^2 x_3^\top x_q x_q^\top x_3 + \frac{4}{D }\frac{v_z^2}{v_y^2}  \delta_1^2 \beta^\top x_1x_1^\top x_q \beta^\top x_2x_2^\top \beta \beta^\top x_3x_3^\top x_q \right)+o\left(\frac{1}{D}\right)\\
    &=& \sigma^2 + \frac{d\sigma^2 }{D} + \frac{1+d}{D} - \frac{(2v_y-v_z)v_z}{D v_y^2} (d-1)\sigma^2 +\frac{4}{D}  \frac{v_xv_z}{v_y^2}  + \frac{6}{D} \frac{v_z^2}{v_y^2} -\frac{2}{D}\frac{v_zv_y}{v_y^2} \sigma_\epsilon^2 \\
    &&+ \frac{2}{D}\frac{v_zv_x}{v_y^2} \sigma_\epsilon^2+\frac{1}{D}\frac{v_z^2}{v_y^2} \sigma_\epsilon^2+\frac{4}{D}\frac{v_z^2}{v_y^2} \sigma_\epsilon^2 +\frac{d}{D}\frac{v_z^2}{v_y^2} \sigma_\epsilon^2+\frac{v_z^2}{v_y^2}\sigma^2\sigma^2_\epsilon +o\left(\frac{1}{D}\right) \\
    &=&   \frac{1+d}{D} +\frac{4}{D}  \frac{v_xv_z}{v_y^2}  + \frac{6}{D} \frac{v_z^2}{v_y^2} 
   +  \frac{3+d}{D} \frac{v_z^2}{v_y^2}\sigma_{\epsilon}^2+\sigma^2 +\frac{d\sigma^2}{D}  - \frac{v_z(2v_x+v_z)}{D v_y^2} (d-1)\sigma^2 +\frac{v_z^2}{v_y^2}\sigma^2\sigma^2_\epsilon +o\left(\frac{1}{D}\right). 
\end{eqnarray*}

\subsection{Proof of Proposition~\ref{prop:perturb}.}\label{proof:prop-perturb}
Because  $L'_{\text{x}}(v_x,v_y,v_z) = O\left(\frac{1}{D}\right)$, $L'_{\text{z}}(v_x,v_y,v_z) = O\left(\frac{1}{D}\right)$ and  $L'_{\text{y}}(v_x,v_y,v_z) = o\left(\frac{1}{D}\right)$. We can conclude that \begin{itemize}
    \item $L'_{\text{x}}(v_x,v_y,v_z) > L'_{\text{y}}(v_x,v_y,v_z)$
    \item $L'_{\text{z}}(v_x,v_y,v_z) > L'_{\text{y}}(v_x,v_y,v_z)$
\end{itemize}

To compare $L'_{\text{z}}(v_x,v_y,v_z)$ and $L'_{\text{x}}(v_x,v_y,v_z)$, we calculate:
\begin{eqnarray*}
    &&  L'_{\text{z}}(v_x,v_y,v_z) -L'_{\text{x}}(v_x,v_y,v_z) \\
     && =  \frac{3+d}{D} \frac{v_z^2}{v_y^2}\sigma_{\epsilon}^2 - \frac{d}{D} \frac{v_x^2}{v_y^2}\sigma_{\epsilon}^2 +\frac{v_z^2-v_x^2}{v_y^2}\sigma^2\sigma^2_\epsilon 
\end{eqnarray*}
Substituting $v_y^*=\frac{(d-1)\sigma^2+2}{(d-1)\sigma^2-2}v_z^*$ and $v_y^* = v_x^* + v_z^*$, we have 
\begin{eqnarray*}
    &&  L'_{\text{z}}(v_x^*,v_y^*,v_z^*) -L'_{\text{x}}(v_x^*,v_y^*,v_z^*) \\
     && =  \frac{3+d}{D} \frac{v_z^2}{v_y^2}\sigma_{\epsilon}^2 - \frac{d}{D} \sigma_{\epsilon}^2 +\frac{v_z^2-v_x^2}{v_y^2}\sigma^2\sigma^2_\epsilon \\
     && = \bigg[  \frac{3+d}{D}  \frac{\left((d-1)\sigma^2-2\right)^2}{\left((d-1)\sigma^2+2\right)^2} - \frac{d}{D}+ \frac{ \left((d-1)\sigma^2-2\right)^2\sigma^2-16\sigma^2}{\left((d-1)\sigma^2+2\right)^2D}\bigg]\sigma^2_\epsilon \\
     && = \frac{\sigma^2_\epsilon}{D}\frac{\left(d-1\right)^2\sigma^6+\left(3d-7\right)\left(d-1\right)\sigma^4-\left(4d+8d^2\right)\sigma^2+12}{\left((d-1)\sigma^2+2\right)^2}
\end{eqnarray*}

Given that the numerator is always positive, we can conclude that when $\left(\left(d-1\right)^2\sigma^6+\left(3d-7\right)\left(d-1\right)\sigma^4-\left(4d+8d^2\right)\sigma^2+12\right)$ is positive, $L'_{\text{z}}(v_x,v_y,v_z) > L'_{\text{x}}(v_x,v_y,v_z)$. Therefore, considering that  $b > a$ are the positive roots of $f(\theta) = \left(d-1\right)^2 \theta^3+\left(3d-7\right)\left(d-1\right)\theta^2-\left(4d+8d^2\right)\theta+12 = 0$. We have:

\begin{itemize}
    \item $L'_{\text{x}}(v_x^*,v_y^*,v_z^*) > L'_{\text{z}}(v_x^*,v_y^*,v_z^*)$ for $\sigma^2 \in \left[ 0, a\right) \cup({b,\infty}) $
    \item $ L'_{\text{x}}(v_x^*,v_y^*,v_z^*) < L'_{\text{z}}(v_x^*,v_y^*,v_z^*)$ for $\sigma^2 \in \left( a, b \right)$.
\end{itemize}


\section{Simulation Setting}\label{appd:sim}
\subsection{Setting of the simulations in Figure~\ref{fig:simulation} and~\ref{fig:simulation2}.}
In the experiments for the simulations in Figure~\ref{fig:simulation} and~\ref{fig:simulation2},  the transformer consists of a single softmax attention layer without read-in or read-out layers. During the training process, we set the learning rate to 1e-4, the batch size to 256, and the number of training steps to 2e5. For the inference process, the batch size was set to 64, with a total of 128,000 inference examples.

\section{More examples for the proposed demonstration format.}
In this section, we provide some additional examples for the proposed demonstration format.

\subsection{Disambiguation QA Dataset}
\small
\begin{tabular}{p{0.99\linewidth}p{0.99\linewidth}}
    \toprule
    \midrule
    Clarify the meaning of sentences with ambiguous pronouns.\\
Q: In the following sentences, explain the antecedent of the pronoun (which thing the pronoun refers to), or state that it is ambiguous. Sentence: The chief told the counselor that they took the day off.\\
Options:\\
(A) The chief took the day off \\  (B) The counselor took the day off \\ (C) Ambiguous\\
Wrong Answer: The incorrect reasoning would choose (B) and state “The psychologist took the day off, and that is why the chief is explaining something to them.” \\
Error: This reasoning is wrong because there is no clear indicator in the sentence that implies the psychologist (counselor) took the day off. \\
Correct Answer:
Let's think step by step. \\
Here we need to determine who the pronoun "they" might be referring to. There are two possible referents for "they", namely the chief and the counselor. The verb "told" might be able to help us determine which one is more likely (if either). Let X be the chief and Y the counselor. The sentence is then of the form "X told Y that (X or Y) did something."\\
Let's consider Y first: "X told Y that Y did something." This case does not make much sense, as Y would already have the information that Y did something, because it is information about themselves.\\
Now, consider X: "X told Y that X did something." This makes sense, because X would be sharing some information about themselves that Y might not have known before.\\
Because in this context, X is the chief and Y is the counselor, the answer should be the chief.\\ So the answer is (A).\\
\\
Q: In the following sentences, explain the antecedent of the pronoun (which thing the pronoun refers to), or state that it is ambiguous. Sentence: The manager sent a message to the secretary, but he didn't reply yet.\\
Options:\\
(A) The secretary didn't reply yet \\ (B) The manager didn't reply yet \\ (C) Ambiguous\\
Wrong Answer: The incorrect reasoning would choose (B) and state “The manager didn’t reply yet, because perhaps the manager is busy or forgot to respond.” \\
Error: This is incorrect because the sentence clearly establishes that the manager was the one who initially sent the message.\\
Correct Answer: Let's think step by step.\\
Here we need to determine who the pronoun "he" might be referring to. There are two possible referents for "he", namely the manager and the secretary. The verbs "sent" and "reply" might be able to help us determine which one is more likely (if either). Let X be the manager and Y the secretary. The sentence is then of the form "X sent a message to Y, but (X or Y) didn't reply yet."\\
Let's consider Y first: "X sent a message to Y, but Y didn't reply yet." This case makes sense, because of the implicit causality of the sentence. Y was the receiver of the message, but Y didn't get back to X yet.
Now, consider X: "X sent a message to Y, but X didn't reply yet." This case doesn't make sense, because X was the initial sender of the message, so it is now Y's turn to write back to X.\\
Because in this context, X is the manager and Y is the secretary, the answer should be the secretary. \\So the answer is (A).\\
\\
Q: In the following sentences, explain the antecedent of the pronoun (which thing the pronoun refers to), or state that it is ambiguous. Sentence: Bailey will plan to meet the director at his office\\
Options:\\
(A) It will be Bailey's office \\ (B) It will be the director's office \\ (C) Ambiguous\\
Wrong Answer: The incorrect reasoning would choose (B) and state "The meeting will take place at the director's office, as it makes sense for a director to conduct meetings in their own office." \\
Error: This is incorrect because the sentence does not explicitly indicate whose office the meeting will take place in.\\
Correct Answer: Let’s think step by step.\\
Here we need to determine who the pronoun "his" might be referring to. There are two possible referents for "his", namely Bailey's and the director's. The verb phrase "plan to meet" might be able to help us determine which one is more likely (if either). Let X be Bailey and Y the director. The sentence is then of the form "X will plan to meet Y at (X or Y)'s office."\\
Let's consider Y first: "X will plan to meet Y at Y's office." This case makes sense, because X might want to meet up with Y at Y's office.\\
Now, consider X: "X will plan to meet Y at X's office." This case also makes sense, because X might want to meet up with Y at X's own office.\\
Because both X and Y are possible at the same time, we conclude that the antecedent of the pronoun is ambiguous.\\ So the answer is (C).\\
    \bottomrule
\end{tabular}
\subsection{Penguins in a Table Dataset}
\small
\begin{tabular}{p{0.99\linewidth}p{0.99\linewidth}}
    \toprule
    \midrule
Answer questions about a table of penguins and their attributes.
\\
Q: Here is a table where the first line is a header and each subsequent line is a penguin:  name, age, height (cm), weight (kg) Louis, 7, 50, 11 Bernard, 5, 80, 13 Vincent, 9, 60, 11 Gwen, 8, 70, 15  For example: the age of Louis is 7, the weight of Gwen is 15 kg, the height of Bernard is 80 cm.  We now add a penguin to the table:\\
James, 12, 90, 12\\
How many penguins are less than 8 years old?\\
Options:\\
(A) 1  \\
(B) 2  \\
(C) 3  \\
(D) 4  \\
(E) 5 \\
Wrong Answer: There are 3 penguins less than 8 years old: Louis, Bernard, and Gwen. So, the answer is (C). \\
Error: This answer incorrectly includes Gwen, who is exactly 8 years old and should not be counted. \\
Correct Answer: \\
A: Let's think step by step. \\
This question focuses on age. We know the following: Louis is 7 years old, Bernard is 5 years old, Vincent is 9 years old, and Gwen is 8 years old. \\
Now, we add James to this table: James is 12 years old. \\
The penguins that are less than 8 years old are Louis and Bernard. \\
There are 2 penguins less than 8 years old. So the answer is (B). \\
\\
Q: Here is a table where the first line is a header and each subsequent line is a penguin:  name, age, height (cm), weight (kg) Louis, 7, 50, 11 Bernard, 5, 80, 13 Vincent, 9, 60, 11 Gwen, 8, 70, 15  For example: the age of Louis is 7, the weight of Gwen is 15 kg, the height of Bernard is 80 cm.  Which is the youngest penguin? \\
Options: \\
(A) Louis \\
(B) Bernard \\
(C) Vincent \\
(D) Gwen\\
(E) James\\ 
Wrong Answer: The youngest penguin is Louis because he is 7 years old. The answer is (A).\\
Error: This answer overlooks the fact that Bernard is 5 years old, making him the youngest.\\
Correct Answer:\\
A: Let's think step by step.\\
This question focuses on age. We know the following: Louis is 7 years old, Bernard is 5 years old, Vincent is 9 years old, and Gwen is 8 years old.\\
According to the table, Bernard (5) is the youngest amongst them.\\
The youngest penguin is Bernard. So the answer is (B).\\
\\
Q: Here is a table where the first line is a header and each subsequent line is a penguin:  name, age, height (cm), weight (kg) Louis, 7, 50, 11 Bernard, 5, 80, 13 Vincent, 9, 60, 11 Gwen, 8, 70, 15  For example: the age of Louis is 7, the weight of Gwen is 15 kg, the height of Bernard is 80 cm.  What is the name of the second penguin sorted by alphabetic order? \\
Options: \\
(A) Louis \\
(B) Bernard \\
(C) Vincent \\
(D) Gwen \\
(E) James \\
Wrong Answer: The names alphabetically are Bernard, Gwen, James, Louis, Vincent. The second penguin is Louis. The answer is (A). \\
Error: This answer incorrectly skips Gwen, which should be the second name alphabetically. \\
Correct Answer: \\
A: Let's think step by step. \\
This question focuses on the name. We know the following: The names of the penguin in the table are Louis, Bernard, Vincent, and Gwen. \\
When we sort their names alphabetically, we get Bernard, Gwen, Louis, Vincent. \\
The name of the second penguin sorted by alphabetical order is Gwen. \\
The name of the second penguin sorted by alphabetic order is Gwen. So the answer is (D). \\
    \bottomrule
\end{tabular}

\subsection{Tracking Shuffled Objects Dataset}
\begin{tabular}{p{0.99\linewidth}p{0.99\linewidth}}
    \toprule
    \midrule
A task requiring determining the final positions of a set of objects given their initial positions and a description of a sequence of swaps.\\
\\
Q: Alice, Bob, and Claire are playing a game. At the start of the game, they are each holding a ball: Alice has a yellow ball, Bob has a blue ball, and Claire has a pink ball.\\
As the game progresses, pairs of players trade balls. First, Claire and Alice swap balls. Then, Alice and Bob swap balls. Finally, Claire and Bob swap balls. At the end of the game, Bob has the\\
Options:\\
(A) yellow ball \quad (B) blue ball \quad
(C) pink ball\\
Wrong Answer: Let's think step by step. \\
(0) At the start: Alice: yellow, Bob: blue, Claire: pink. \\
(1) Claire and Alice swap balls: Alice: pink, Bob: blue, Claire: yellow. \\
(2) Alice and Bob swap balls: Alice: blue, Bob: pink, Claire: yellow. \\
At the end of the game, Bob has the pink ball. So the answer is (C). \\
Error: This answer forgets the last swap between Claire and Bob, incorrectly concluding that Bob holds the pink ball from the second swap, when in fact, Bob trades with Claire again.\\
Correct answer:\\
A: Let's think step by step.\\
(0) At the start: Alice: yellow, Bob: blue, Claire: pink.\\
(1) Claire and Alice swap balls: Alice: pink, Bob: blue, Claire: yellow.\\
(2) Alice and Bob swap balls: Alice: blue, Bob: pink, Claire: yellow.\\
(3) Claire and Bob swap balls: Alice: blue, Bob: yellow, Claire: pink.\\
At the end of the game, Bob has the yellow ball. So the answer is (A).\\
\\
Q: Alice, Bob, and Claire are playing a game. At the start of the game, they are each holding a ball: Alice has a white ball, Bob has a purple ball, and Claire has a pink ball.
As the game progresses, pairs of players trade balls. First, Bob and Alice swap balls. Then, Bob and Claire swap balls. Finally, Bob and Alice swap balls. At the end of the game, Alice has the\\
Options:\\
(A) white ball\quad
(B) purple ball\quad
(C) pink ball\\
Wrong Answer: Let's think step by step. \\
(0) At the start: Alice: white, Bob: purple, Claire: pink. \\
(1) Bob and Alice swap balls: Alice: purple, Bob: white, Claire: pink. \\
(2) Bob and Alice swap balls: Alice: white, Bob: pink, Claire: pink.\\
At the end of the game, Bob Alice has the white ball. So the answer is (A).\\
Error: This answer assumes Alice gets her original ball back, ignoring the correct sequence of swaps where Alice ends up with the pink ball after the final exchange.\\
Correct answer:\\
A: Let's think step by step.\\
(0) At the start: Alice: white, Bob: purple, Claire: pink.\\
(1) Bob and Alice swap balls: Alice: purple, Bob: white, Claire: pink.\\
(2) Bob and Claire swap balls: Alice: purple, Bob: pink, Claire: white.\\
(3) Bob and Alice swap balls: Alice: pink, Bob: purple, Claire: white.\\
At the end of the game, Alice has the pink ball. So the answer is (C).\\
\\
Q: Alice, Bob, and Claire are dancers at a square dance. At the start of a song, they each have a partner: Alice is dancing with Lola, Bob is dancing with Rodrigo, and Claire is dancing with Patrick.
Throughout the song, the dancers often trade partners. First, Alice and Bob switch partners. Then, Claire and Bob switch partners. Finally, Bob and Alice switch partners. At the end of the dance, Alice is dancing with\\
Options:\\
(A) Lola\quad
(B) Rodrigo\quad
(C) Patrick\\
Wrong Answer: Let's think step by step. \\
(0) At the start: Alice: Lola, Bob: Rodrigo, Claire: Patrick.\\
Error: This answer mistakenly assumes that Alice ends up with her original partner, overlooking the fact that Alice has switch partners for twice.\\
Correct answer:\\
A: Let's think step by step.\\
(0) At the start: Alice: Lola, Bob: Rodrigo, Claire: Patrick.\\
(1) Alice and Bob switch partners: Alice: Rodrigo, Bob: Lola, Claire: Patrick.\\
(2) Claire and Bob switch partners: Alice: Rodrigo, Bob: Patrick, Claire: Lola.\\
(3) Bob and Alice switch partners: Alice: Patrick, Bob: Rodrigo, Claire: Lola.\\
At the end of the dance, Alice is dancing with Patrick. So the answer is (C).\\
\bottomrule
\end{tabular}

\subsection{Date Understanding Dataset}
\begin{tabular}{p{0.99\linewidth}p{0.99\linewidth}}
    \toprule
    \midrule
    Infer the date from context.\\
\\
Q: Today is Christmas Eve of 1937. What is the date 10 days ago in MM/DD/YYYY?\\
Options:\\
(A) 12/14/2026\\
(B) 12/14/1950\\
(C) 12/14/2007\\
(D) 12/14/1937\\
(E) 07/14/1938\\
(F) 12/14/1988\\
Wrong Answer: The date 10 days ago from Christmas Eve of 1937 would be 12/14/1950.\\
Error: This answer incorrectly chooses a future year. The correct answer is 12/14/1937, as 10 days before Christmas Eve of 1937 is December 14, 1937.\\
Correct Answer:\\
Let's think step by step.\\
If today is Christmas Eve of 1937, then today's date is December 24, 1937. 10 days before today is December 14, 1937, that is 12/14/1937. So the answer is (D).\\
\\
Q: Tomorrow is 11/12/2019. What is the date one year ago from today in MM/DD/YYYY?\\
Options:\\
(A) 09/04/2018\\
(B) 11/11/2018\\
(C) 08/25/2018\\
(D) 11/02/2018\\
(E) 11/04/2018\\
Wrong Answer: The date one year ago from today would be 09/04/2018.\\
Error: This answer selects an incorrect month and day. Since today is 11/11/2019, one year ago would be 11/11/2018, making the correct answer 11/11/2018.\\
Correct Answer:\\
Let's think step by step.\\
If tomorrow is 11/12/2019, then today is 11/11/2019. The date one year ago from today is 11/11/2018. So the answer is (B).\\
\\
Q: Jane and John married on Jan 2, 1958. It is their 5-year anniversary today. What is the date tomorrow in MM/DD/YYYY?\\
Options:\\
(A) 01/11/1961\\
(B) 01/03/1963\\
(C) 01/18/1961\\
(D) 10/14/1960\\
(E) 01/03/1982\\
(F) 12/03/1960\\
Wrong Answer: Since it is their 5-year anniversary today, the date tomorrow would be 01/18/1961.\\
Error: This answer incorrectly calculates the year. Since Jane and John were married on January 2, 1958, and today is their 5-year anniversary, the correct date tomorrow would be 01/03/1963.\\
Correct Answer:\\
Let's think step by step.\\
If Jane and John married on Jan 2, 1958, then and if it is their 5-year anniversary today, then today's date is Jan 2, 1963. The date tomorrow is Jan 3, 1963, that is 01/03/1963. So the answer is (B).\\
    \bottomrule
\end{tabular}

\subsection{GSM8K Dataset}

\begin{tabular}{p{0.99\linewidth}p{0.99\linewidth}}
    \toprule
    \midrule
    Question: There are 15 trees in the grove. Grove workers will plant trees in the grove today. After they are done, there will be 21 trees. How many trees did the grove workers plant today?\\
Wrong Answer: The incorrect reasoning might be, "There were 15 trees, and there will be 21 after planting. Adding these gives 15 + 21 = 36 trees planted today."\\
Error: This is wrong because the correct method is to subtract the initial 15 trees from the final 21. The correct answer is 6 trees planted today.\\
Correct Answer:\\
Let's think step by step\\
There are 15 trees originally.\\
Then there were 21 trees after some more were planted.\\
So there must have been 21 - 15 = 6.\\
The answer is 6.\\
\\
Question: If there are 3 cars in the parking lot and 2 more cars arrive, how many cars are in the parking lot?
Wrong Answer: The incorrect reasoning might be, "There are 3 cars, and 2 more arrive. Multiplying 3 by 2 gives 6, so there are 6 cars."\\
Error: This is wrong because the correct operation is addition, not multiplication. You should add the 2 new cars to the 3 already there, giving 5 cars.\\
Correct Answer:\\
Let's think step by step\\
There are originally 3 cars.\\
2 more cars arrive.\\
3 + 2 = 5.\\
The answer is 5.\\
\\
Question: Leah had 32 chocolates and her sister had 42. If they ate 35, how many pieces do they have left in total?\\
Wrong Answer: The incorrect reasoning might be, "Leah had 32 chocolates and her sister had 42. Subtracting the 35 they ate from 42 gives 7 chocolates left."\\
Error: This is incorrect because the total pieces they had initially is 32 + 42 = 74, and subtracting 35 from 74 gives the correct answer of 39 pieces left.\\
Correct Answer:\\
Let's think step by step\\
Originally, Leah had 32 chocolates.\\
Her sister had 42.\\
So in total they had 32 + 42 = 74.\\
After eating 35, they had 74 - 35 = 39.\\
The answer is 39.\\
\\
Question: Jason had 20 lollipops. He gave Denny some lollipops. Now Jason has 12 lollipops. How many lollipops did Jason give to Denny?\\
Wrong Answer: The incorrect reasoning might be, "Jason had 20 lollipops, and now he has 12. Adding 12 to 20 gives 32 lollipops given to Denny."\\
Error: This is wrong because the correct operation is subtraction, not addition. Subtracting 12 from 20 gives the correct answer of 8 lollipops given to Denny.\\
Correct Answer:\\
Let's think step by step\\
Jason started with 20 lollipops.\\
Then he had 12 after giving some to Denny.\\
So he gave Denny 20 - 12 = 8.\\
The answer is 8.\\
    \bottomrule
\end{tabular}

\section{Details about the computing infrastructure.}
\textbf{Experiments for the simulations in Figure~\ref{fig:simulation} and~\ref{fig:simulation2}.}
We conducted the experiments on a single A6000 GPU with 48GB of memory. Training the transformer took approximately 30 minutes, while inference took around 40 seconds. 

\textbf{Experiments in Section~\ref{sec:exp}}. For these experiments, we called all models directly using their APIs, except for DeepSeek 67, for which we downloaded the checkpoint and ran the model on four A6000 GPUs. For the BBH benchmark datasets (each containing 250 examples), it took approximately 3–6 hours to generate solutions for all questions. For the GSM8k dataset (which contains 1,319 examples), the total generation time was around 14 hours.
\end{document}